%% file: main.tex
\documentclass[runningheads]{llncs}
\usepackage{graphicx}
\usepackage{comment}
\usepackage{amsmath,amssymb} 
\usepackage{color}
\usepackage{adjustbox}
\usepackage{caption}
\usepackage{multirow}
\usepackage[breaklinks=true]{hyperref}
\usepackage{breakcites}

\input{macros}

\usepackage{amssymb}
\usepackage{pifont}
\newcommand{\cmark}{\ding{51}}%
\newcommand{\xmark}{\ding{55}}%

\makeatletter
\newcommand{\printfnsymbol}[1]{%
\textsuperscript{\@fnsymbol{#1}}%
}

\begin{document}
\pagestyle{headings}
\mainmatter
\def\ECCVSubNumber{1561}  

\title{H3DNet: 3D Object Detection Using Hybrid Geometric Primitives} 

\titlerunning{H3DNet}
%
\author{Zaiwei Zhang\inst{1} \and
Bo Sun\thanks{Equal Contribution}\inst{1} \and
Haitao Yang\printfnsymbol{1}\inst{1} \and
Qixing Huang\inst{1}
}
\authorrunning{Zhang et al.}
%
\institute{The University of Texas at Austin, Austin, Texas, USA, 78710}
\maketitle

\begin{abstract}
We introduce H3DNet, which takes a colorless 3D point cloud as input and outputs a collection of oriented object bounding boxes (or BB) and their semantic labels. The critical idea of H3DNet is to predict a hybrid set of geometric primitives, i.e., BB centers, BB face centers, and BB edge centers. We show how to convert the predicted geometric primitives into object proposals by defining a distance function between an object and the geometric primitives. This distance function enables continuous optimization of object proposals, and its local minimums provide high-fidelity object proposals. H3DNet then utilizes a matching and refinement module to classify object proposals into detected objects and fine-tune the geometric parameters of the detected objects. The hybrid set of geometric primitives not only provides more accurate signals for object detection than using a single type of geometric primitives, but it also provides an overcomplete set of constraints on the resulting 3D layout. Therefore, H3DNet can tolerate outliers in predicted geometric primitives. Our model achieves state-of-the-art 3D detection results on two large datasets with real 3D scans, ScanNet and SUN RGB-D. \textcolor{red}{Our code is open-sourced at} \href{https://github.com/zaiweizhang/H3DNet}{\textcolor{red}{here}}.
\end{abstract}

\input{01_intro}

\input{02_related_works}

\input{03_overview}
\input{04_primitive_module}

\input{05_proposal_module}

\input{06_classify_and_refine_module}
\input{07_training_details}
\input{08_results}
\input{09_conclusions}

%
%

\input{main.bbl}
\newpage
\clearpage
\appendix

\input{100_supp_intro}
\input{10_supp_proof}
\input{11_supp_network}
\input{12_supp_primitive}
\input{13_supp_result}

\end{document}

%% file: macros.tex
\newcommand{\R}{\mathbb{R}}
\let \bs=\mathbf
\let \set=\mathcal

\def \gt {\textup{gt}}

\def \gt {\textup{gt}}

\def \saliency {\textup{\saliency}}

\def \gt {\mathit{gt}}

\def \path {\mathit{path}}

\def \line {\textup{edge}}

\def \cent {\textup{center}}
\def \face {\textup{face}}
\def \edge {\textup{edge}}
\def \cbox {\textup{box}}

\let \set = \mathcal
\let \bs = \boldsymbol

%% file: 01_intro.tex
\section{Introduction}
\label{Section:Intro}

Object detection is a fundamental problem in visual recognition. In this work, we aim to detect the 3D layout (i.e., oriented 3D bounding boxes (or BBs) and associated semantic labels) from a colorless 3D point cloud. This problem is fundamentally challenging because of the irregular input and a varying number of objects across different scenes. Choosing suitable intermediate representations to integrate low-level object cues into detected objects is key to the performance of the resulting system. While early works~\cite{Song2014SlidingSF,Song_2016_CVPR} classify sliding windows for object detection, recent works~\cite{Ren_2016_CVPR,chen2017multi,LahoudG17,qi2018frustum,zhou2018voxelnet,qi2019votenet,yi2019gspn,shi2019pointrcnn,yi2019gspn,pham2019jsis3d,wang2019associatively} have shown the great promise of designing end-to-end neural networks to generate, classify, and refine object proposals. 

This paper introduces H3DNet, an end-to-end neural network that utilizes a novel intermediate representation for 3D object detection. Specifically, H3DNet first predicts a hybrid and overcomplete set of geometric primitives (i.e., BB centers, BB face centers, and BB edge centers) and then detects objects to fit these primitives and their associated features. This regression methodology, which is motivated from the recent success of keypoint-based pose regression for 6D object pose estimation~\cite{LepetitMF09,PavlakosZCDD17,KendallC17,LiWJXF18,PengLHZB19,HybridPose6D}, displays two appealing advantages for 3D object detection. First, each type of geometric primitives focuses on different regions of the input point cloud (e.g., points of an entire object for predicting the BB center and points of a planar boundary surface for predicting the corresponding BB face center). Combing diverse primitive types can add the strengths of their generalization behaviors. On new instances, they offer more useful constraints and features than merely using one type of primitives. Second, having an overcomplete set of primitive constraints can tolerate outliers in predicted primitives (e.g., using robust functions) and reduce the influence of individual prediction errors. The design of H3DNet fully practices these two advantages. 

Specifically, H3DNet consists of three modules. The first module computes dense pointwise descriptors and uses them to predict geometric primitives and their latent features. The second module converts these geometric primitives into object proposals. A key innovation of H3DNet is to define a parametric distance function that evaluates the distance between an object BB and the predicted primitives. This distance function can easily incorporate diverse and overcomplete geometric primitives. Its local minimums naturally correspond to object proposals. This method allows us to optimize object BBs continuously and generate high-quality object proposals from imprecise initial proposals. 

\begin{figure}[t]
\centering
\includegraphics[width=1.0\columnwidth]{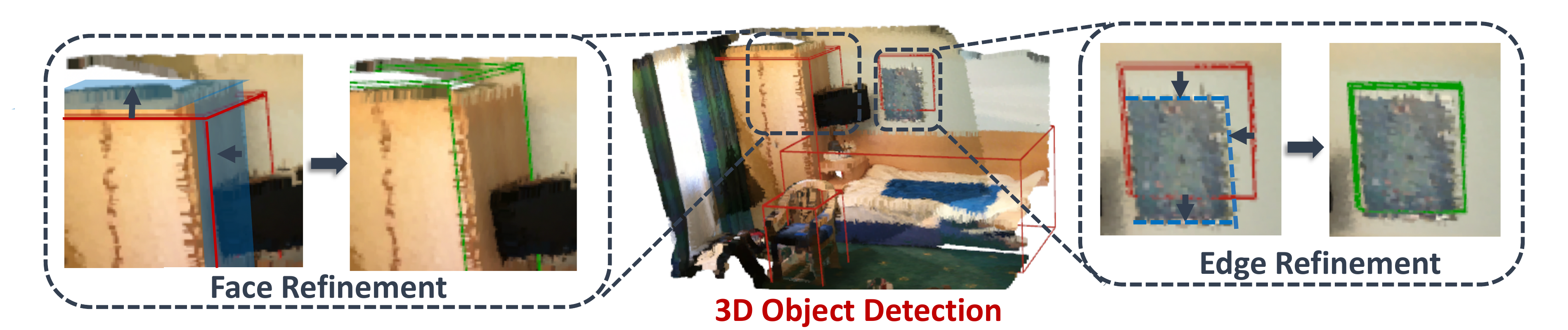}
\caption{\small{Our approach leverages a hybrid and overcomplete set of geometric primitives to detect and refine 3D object bounding boxes (BBs). Note that red BBs are initial object proposals, green BBs are refined object proposals, and blue surfaces and lines are hybrid geometric primitives. }}
\label{Figure:Teaser}
\end{figure}

The last module of H3DNet classifies each object proposal as a detected object or not, and also predicts for each detected object an offset vector of its geometric parameters and a semantic label to fine-tune the detection result. The performance of this module depends on the input. As each object proposal is associated with diverse geometric primitives, H3DNet aggregates latent features associated with these primitives, which may contain complementary semantic and geometric information, as the input to this module. We also introduce a network design that can handle a varying number of geometric primitives. 

We have evaluated H3DNet on two popular benchmark datasets ScanNet and SUN RGB-D. On ScanNet, H3DNet achieved 67.2\% in mAP (0.25), which corresponded to a 8.5\% relative improvement from state-of-the-art methods that merely take the 3D point positions as input. On SUN RGB-D, H3DNet achieved 60.1\% in mAP (0.25), which corresponded to a 2.4\% relative improvement from the same set of state-of-the-art methods. Moreover, on difficult categories of both datasets (i.e., those with low mAP scores), the performance gains of H3DNet are significant (e.g., from 38.1/47.3/57.1 to 51.9/61.0/75.3/ on window/door/shower-curtain, respectively). We have also performed an ablation study on H3DNet. Experimental results justify the importance of regressing a hybrid and overcomplete set of geometric primitives for generating object proposals and aggregating features associated with matching primitives for classifying and refining detected objects. In summary, the contributions of our work are:
\begin{itemize}
\item Formulation of object detection as regressing and aggregating an overcomplete set of geometric primitives
\item Predicting multiple types of geometric primitives that are suitable for different object types and scenes 
\item State-of-the-art results on SUN RGB-D and ScanNet with only point clouds
\end{itemize}

%% file: 02_related_works.tex
\section{Related Works}
\label{Section:Related:Works}

\noindent\textbf{3D object detection.} From the methodology perspective, there are strong connections between 3D object detection approaches and their 2D counterparts. Most existing works follow the approach of classifying candidate objects that are generated using a sliding window~\cite{Song2014SlidingSF,Song_2016_CVPR} or more advanced techniques~\cite{qi2018frustum,zhou2018voxelnet,qi2019votenet,yi2019gspn,shi2019pointrcnn,yi2019gspn,pham2019jsis3d,wang2019associatively}. Objectness classification involves template-based approaches or deep neural networks. The key differences between 2D approaches and 3D approaches lie in feature representations. For example, \cite{LinFU13} leverages a pair-wise semantic context potential to guide the proposals' objectness score. \cite{Ren_2016_CVPR} uses clouds of oriented gradients (COG) for object detection. \cite{Hou_2019_CVPR_3D-SIS} utilizes the power of 3D convolution neural networks to identify locations and keypoints of 3D objects. Due to the computational cost in the 3D domain, many methods utilize 2D-3D projection techniques to integrate 2D object detection and 3D data processing. For example, MV3D~\cite{chen2017multi} and VoxelNet~\cite{zhou2018voxelnet} represent the 3D input data in a bird’s-eye view before proceeding to the rest of the pipeline. Similarly, \cite{KimXS13,LahoudG17,qi2018frustum} first process 2D inputs to identify candidate 3D object proposals.

Point clouds have emerged as a powerful representation for 3D deep learning, particularly for extracting salient geometric features and spatial locations (c.f.~\cite{qi2017pointnet,qi2017pointnet++}). Prior usages of point-based neural networks include classification~\cite{qi2017pointnet,qi2017pointnet++,li2018pointcnn,wang2017cnn,klokov2017escape,xu2018spidercnn,wang2018dynamic,xie2018attentional,hermosilla2018monte,hua2018pointwise}, segmentation~\cite{qi2017pointnet++,su2018splatnet,atzmon2018point,li2018pointcnn,wang2017cnn,tatarchenko2018tangent,klokov2017escape,xu2018spidercnn,wang2018dynamic,xie2018attentional,hermosilla2018monte,hua2018pointwise,graham20183d,wang2019associatively}, normal estimation~\cite{atzmon2018point}, and 3D reconstruction~\cite{tatarchenko2017octree,fan2017point,yang2018foldingnet}.

There are also growing interests in object detection from point clouds~\cite{qi2019votenet,yi2019gspn,shi2019pointrcnn,yi2019gspn,pham2019jsis3d,wang2019associatively}. H3DNet is most relevant to~\cite{qi2019votenet}, which leverages deep neural networks to predict object bounding boxes. The key innovation of H3DNet is that it utilizes an overcomplete set of geometric primitives and a distance function to integrate them for object detection. This strategy can tolerate inaccurate primitive predictions (e.g., due to partial inputs).   


\noindent\textbf{Multi-task 3D understanding.} Jointly predicting different types of geometric primitives is related to multi-task learning~\cite{baxter1997bayesian,kendall2018multi,sener2018multi,DBLP:journals/corr/Ruder17a,DBLP:journals/corr/ZhangY17aa,luvizon20182d,liang2019multi,lahoud20193d,pham2019jsis3d,zou2018df,Zhang_2019_CVPR}, where incorporating multiple relevant tasks together boosts the performance of feature learning. In a recent work HybridPose~\cite{HybridPose6D}, Song et al. show that predicting keypoints, edges between keypoints, and symmetric correspondences jointly lift the prediction accuracies of each type of features. In this paper, we show that predicting BB centers, BB face centers, and BB edge centers together help to improve the generalization behavior of primitive predictions. 

\noindent\textbf{Overcomplete constraints regression.} The main idea of H3DNet is to incorporate an overcomplete set of constraints. This approach achieves considerable performance gains from~\cite{qi2019votenet}, which uses a single type of geometric primitives. At a conceptual level, similar strategies have been used in tasks of object tracking~\cite{wang2018multi}, zero-shot fine-grained classification~\cite{akata2016multi}, 6D object pose estimation~\cite{HybridPose6D} and relative pose estimation between scans~\cite{zhenpei20}, among others. Compared to these works, the novelties of H3DNet lie in designing hybrid constraints that are suitable for object detection, continuous optimization of object proposals, aggregating hybrid features for classifying and fine-tuning object proposals, and end-to-end training of the entire network.

%% file: 03_overview.tex
\section{Approach}
\label{Section:Approach}

This section describes the technical details of H3DNet. Section~\ref{Section:Overview} presents an approach overview. Section~\ref{Section:Constraint:Module} to Section~\ref{Section:Network:Training} elaborate on the network design and the training procedure of H3DNet. 

\subsection{Approach Overview}
\label{Section:Overview}

\begin{figure*}[t!]
 \includegraphics[width=1.0\textwidth]{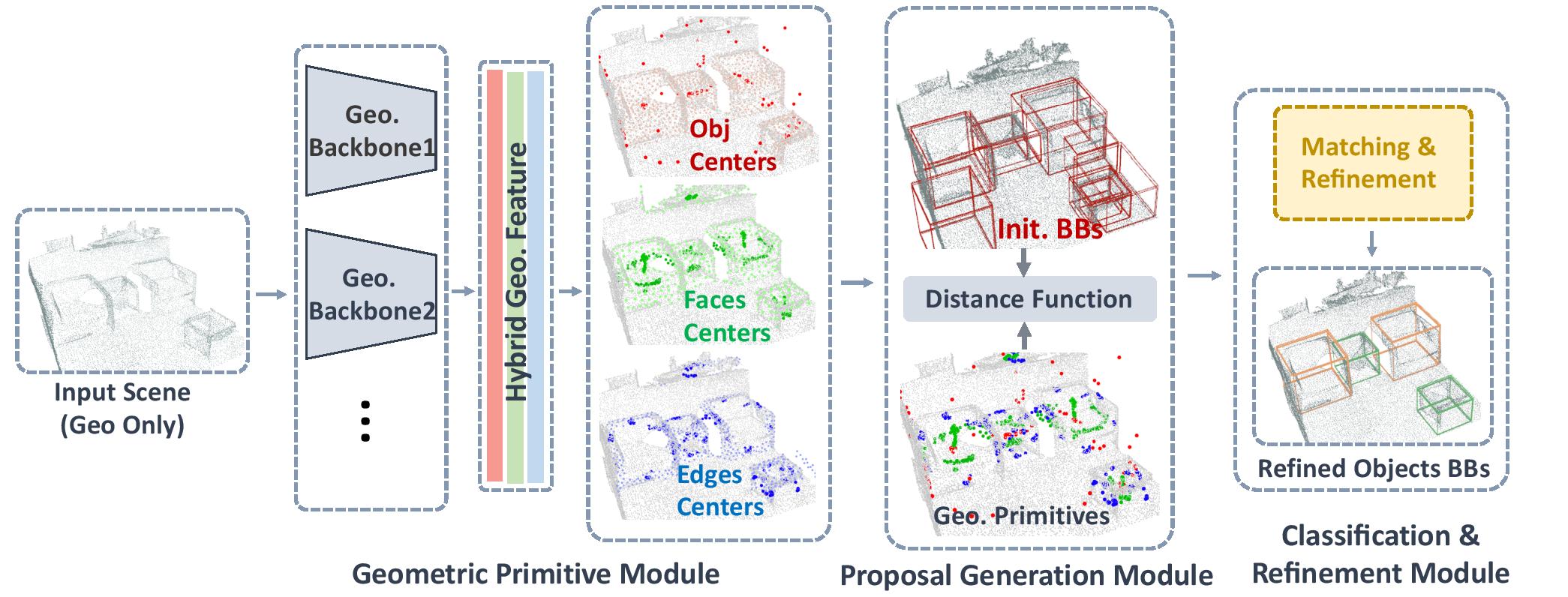}
\caption{\small{H3DNet consists of three modules. The first module computes a dense descriptor and predicts three geometric primitives, namely, BB centers, BB face centers, and BB edge centers. The second module converts geometric primitives into object proposals. The third module classifies object proposals and refines the detected objects.}}
\label{Figure:Approach:Overview}
\end{figure*}

As illustrated in Figure~\ref{Figure:Approach:Overview}, the input of H3DNet is a dense set of 3D points (i.e., a point cloud) $S \in \R^{3\times n}$ ($n = 40000$ in this paper). Such an input typically comes from depth sensors or the result of multi-view stereo matching. The output is given by a collection of (oriented) bounding boxes (or BB) $\set{O}_S\in \overline{\set{O}}$, where $\overline{\set{O}}$ denotes the space of all possible objects. Each object $o \in \overline{\set{O}}$ is given by its class label $l_o \in \set{C}$, where $\set{C}$ is pre-defined, its center $\bs{c}_o = (c_o^x, c_o^y, c_o^z)^T\in \R^3$ in a world coordinate system, its scales $\bs{s}_o = (s_o^x, s_o^y, s_o^z)^T\in \R^3$, and its orientation $\bs{n}_o = (\bs{n}_o^x, \bs{n}_o^y)^T \in \R^2$ in the xy-plane of the same world coordinate system (note that the upright direction of an object is always along the z axis). 

H3DNet consists of three modules, starting from geometric primitive prediction, to proposal generation, to object refinement. The theme is to predict and integrate an overcomplete set of geometric primitives, i.e., BB centers, BB face centers, and BB edge centers. The entire network is trained end-to-end. 

\noindent\textbf{Geometric primitive module.}
The first module of H3DNet takes a point cloud $S$ as input and outputs a set of geometric primitives $\set{P}_S$ that predicts locations of BB centers, BB face centers, and BB edge centers of the underlying objects. The network design extends that of~\cite{qi2019votenet}. Specifically, it combines a sub-module for extracting dense point-wise descriptors and sub-modules that take point-wise descriptors as input and output offset vectors between input points and the corresponding centers. The resulting primitives are obtained through clustering. In addition to locations, each predicted geometric primitive also possesses a latent feature that is passed through subsequent modules of H3DNet. 

In contrast to~\cite{qi2019votenet}, H3DNet exhibits two advantages. First, since only a subset of predicted geometric primitives is sufficient for object detection, the detected objects are insensitive to erroneous predictions. Second, different types of geometric primitives show complementary strength. For example, BB centers are accurate for complete and solid objects, while BB face centers are suitable for partial objects that possess rich planar structures. 

\noindent\textbf{Proposal generation module.} The second module takes predicted geometric primitives as input and outputs a set of object proposals. A critical innovation of H3DNet is to formulate object proposals as local minimums of a distance function. 
This methodology is quite flexible in several ways. First, it is easy to incorporate overcomplete geometric primitives, each of which corresponds to an objective term of the distance function. Second, it can handle outlier predictions and mispredictions using robust norms. Finally, it becomes possible to optimize object proposals continuously, and this property relaxes the burden of generating high-quality initial proposals. 

\noindent\textbf{Classification and refinement module.} The last module of H3DNet classifies each object proposal into a detected object or not. This module also computes offset vectors to refine the BB center, BB size, and BB orientation of each detected object, and a semantic label. The key idea of this module is to aggregate features of the geometric primitives that are close to the corresponding primitives of each object proposal. Such aggregated features carry rich semantic information that is unavailable in the feature associated with each geometric primitive. 

%% file: 04_primitive_module.tex
\subsection{Primitive Module}
\label{Section:Constraint:Module}

The first module of H3DNet predicts a set of geometric primitives from the input point cloud. Each geometric primitive provides some constraints on the detected objects. In contrast to most prior works that compute a minimum set of primitives, i.e., that is sufficient to determine the object bounding boxes, H3DNet leverages an overcomplete set of geometric primitives, i.e., BB centers, BB face centers, and BB edge centers. In other words, these geometric primitives can provide up-to 19 positional constraints for one BB. As we will see later, they offer great flexibilities in generating, classifying, and refining object proposals. 


Similar to~\cite{qi2019votenet}, the design of this module combines a descriptor sub-module and a prediction sub-module. The descriptor sub-module computes dense point-wise descriptors. Its output is fed into the prediction sub-module, which consists of three prediction branches. Each branch predicts one type of geometric primitives. Below we provide the technical details of the network design.


\noindent\textbf{Descriptor sub-module.} The output of the descriptor sub-module provides semantic information to group points for predicting geometric primitives (e.g., points of the same object for BB centers and points of the same planar boundary faces for BB face centers). Instead of using a single descriptor computation tower~\cite{qi2019votenet}, H3DNet integrates four separate descriptor computation towers. The resulting descriptors are concatenated together for primitive prediction and subsequent modules of H3DNet. Our experiments indicate that this network design can learn distinctive features for predicting each type of primitives. However, it does not lead to a significant increase in network complexity. 


\noindent\textbf{BB center prediction.} The same as~\cite{qi2019votenet}, H3DNet leverages a network with three fully connected layers to predict the offset vector between each point and its corresponding object center. The resulting BB centers are obtained through clustering (c.f.~\cite{qi2019votenet}). Note that in additional to offset vectors, H3DNet also computes an associated feature descriptor for each BB center. These feature descriptors serve as input feature representations for subsequent modules of H3DNet. 

Predictions of BB centers are accurate on complete and rectangular shaped objects. However, there are shifting errors for partial and/or occluded objects, and thin objects, such as pictures or curtains, due to imbalanced weighting for offset prediction. This motivates us to consider centers of BB faces and BB edges. 

\noindent\textbf{BB face center prediction.} Planar surfaces are ubiquitous in man-made scenes and objects. Similar to BB center, H3DNet uses 3 fully connected layers to perform point-wise predictions. The predicted attributes include a flag that indicates whether a point is close to a BB face or not and if so, an offset vector between that point and its corresponding BB face center. For training, we generate the ground-truth labels by computing the closest BB face for each point. We say a point lies close to a BB face (i.e., a positive instance) if that distance is smaller than $0.2m$. Similar to BB centers, each BB face center prediction also possesses a latent feature descriptor that is fed into the subsequent modules. 

Since face center predictions are only affected by points that are close to that face, we found that they are particularly useful for objects with rich planar patches (e.g., refrigerator and shower-curtain) and incomplete objects.

\noindent\textbf{BB edge center prediction.} Boundary line features form another type of geometric cues in all 3D scenes and objects. Similar to BB faces, H3DNet employs 3 fully connected layers to predict for each point a flag that indicates whether it is close to a BB edge or not and if so, an offset vector between that point and the corresponding BB edge center. 
The same as BB face centers, we generate ground-truth labels by computing the closest BB edge for each point. We say a point lies close to a BB edge if the closest distance is smaller than $0.2m$. Again, each BB edge center prediction possesses a latent feature of the same dimension. Compared to BB centers and BB face centers, BB edge centers are useful for objects where point densities are irregular (e.g., with large holes) but BB edges appear to be complete (e.g., window and computer desk).

As analyzed in details in the supplemental material, error distributions of different primitives are largely uncorrelated with each other. Such uncorrelated prediction errors provide a foundation for performance boosting when integrating them together for detecting objects.

%% file: 05_proposal_module.tex
\subsection{Proposal Module}
\label{Section:Proposal:Module}
%

After predicting geometric primitives, H3DNet proceeds to compute object proposals. Since the predicted geometric primitives are overcomplete, H3DNet converts them into a distance function and generates object proposals as local minimums of this distance function. This approach, which is the crucial contribution of H3DNet, exhibits several appealing properties. First, it automatically incorporates multiple geometric primitives to determine the parameters of each object proposal. Second, the distance function can optimize object proposals continuously. The resulting local minimums are insensitive to initial proposals, allowing us to use simple initial proposal generators. Finally, each local minimum is attached to different types of geometric primitives, which carry potentially complementary semantic information. As discussed in Section~\ref{Subsec:Refinement:Module}, the last module of H3DNet builds upon this property to classify and refine object proposals. 

\noindent\textbf{Proposal distance function.} The proposal distance function $F_{S}(o)$ measures a cumulative proximity score between the predicted geometric primitives $\set{P}_{S}$ and the corresponding object primitives of an object proposal $o$. Recall that $l_o \in \set{C}$, $\bs{c}_o \in \R^3$, $\bs{s}_o \in \R^3$, and $\bs{n}_o \in \R^2$ denote the label, center, scales, and orientation of $o$. With $\bs{o} = (\bs{c}_o^T,\bs{s}_o^T, \bs{n}_o^T)^T$ we collect all the geometric parameters of $o$. Note that each object proposal $o$ has 19 object primitives (i.e., one BB center, six BB face centers, and twelve BB edge centers). Let $\bs{p}_i(o), 1\leq i \leq 19$ be the location of the $i$-th primitive of $o$. Denote $t_i \in \set{T}:=\{\cent, \face, \edge\}$ as the type of the $i$-th primitive. Let $\set{P}_{t,S}\subseteq \set{P}_{S}$ collect all predicted primitives with type $t \in \set{T}$. We define
\begin{equation}
F_{S}(o):= \sum\limits_{t\in \set{T}}\beta_t \sum\limits_{\bs{c}\in \set{P}_{t,S}} \min\Big(\min\limits_{1\leq i \leq 19,t_i=t}\|\bs{c}_i-\bs{p}_i(o)\|^2-\delta,0\Big).
\label{Eq:Potential:Func}
\end{equation}
In other words, we employ the truncated L2-norm to match predicted primitives and closest object primitives. $\beta_t$ describes the trade-off parameter for type $t$. Both $\beta_t$ and the truncation threshold $\delta$ are determined via cross-validation.

\noindent\textbf{Initial proposals.} H3DNet detects object proposals by exploring the local minimums of the distance function from a set of initial proposals. From the perspective of optimization, we obtain the same local minimum from any initial solution that is sufficiently close to that local minimum. This means the initial proposals do not need to be exact. In our experiments, we found that a simple object proposal generation approach is sufficient. Specifically, H3DNet utilizes the method of~\cite{qi2019votenet}, which initializes an object proposal from each detected BB center. 


\noindent\textbf{Proposal refinement.} By minimizing $F_{S}$, we refine each initial proposal. Note that different initial proposals may share the same local minimum. The final object proposals only collect distinctive local minimums.

%% file: 06_classify_and_refine_module.tex
\begin{figure*}[t!]
\includegraphics[width=1.0\textwidth]{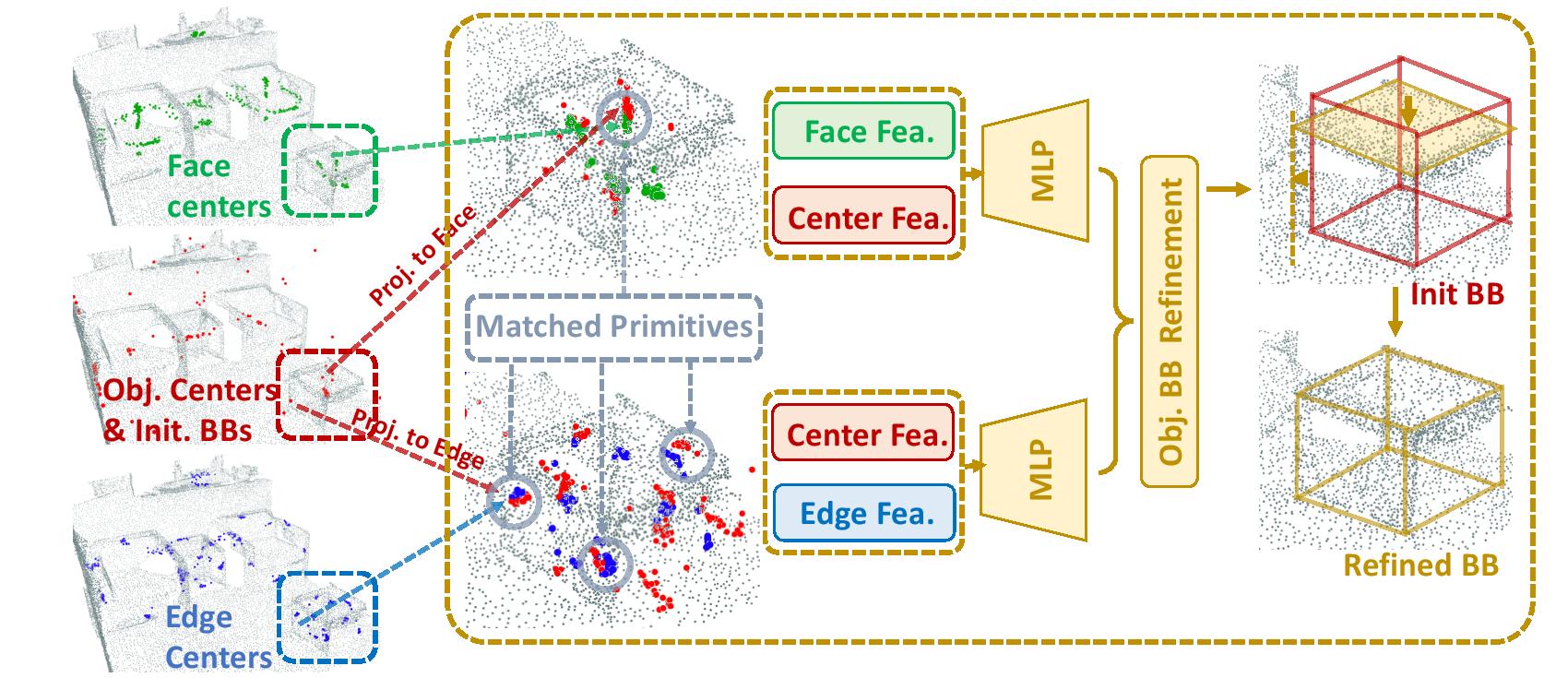}
\caption{\small{Illustration of the matching, feature aggregation and refinement process.}}
\label{Figure:Geometric:Constraints}
\end{figure*}

\subsection{Classification and Refinement Module}
\label{Subsec:Refinement:Module}

The last module of H3DNet takes the output of the proposal module as input and outputs a collection of detected objects. This module combines a classification sub-module and a refinement sub-module. The classification sub-module determines whether each object proposal is an object or not. The refinement sub-module predicts for each detected object the offsets in BB center, BB size, and BB orientation and a semantic label. 

The main idea is to aggregate features associated the primitives (i.e., object centers, edge centers, and face centers) of each object proposal. Such features capture potentially complementary information, yet only at this stage (i.e., after we have detected groups of matching primitives) it becomes possible to fuse them together to determine and fine-tune the detected objects.

As illustrated in Figure \ref{Figure:Geometric:Constraints}, we implement this sub-module by combing four fully connected layers. The input layer concatenates input features of 19 object primitives of an object proposal (i.e., one BB center, six BB face centers, and twelve BB edge centers). Each input feature integrates features associated with primitives that are in the neighborhood of the corresponding object primitive. To address the issue that there is a varying number of neighborhood primitives (e.g., none or multiple), we utilize a variant of the max-pooling layer in PointNet~\cite{qi2017pointnet,qi2017pointnet++} to compute the input feature. Specifically, the input to each max-pooling layer consists of the feature associated with the input object proposal, which addresses the issue of no matching primitives, and 32 feature points that are randomly sampled in the neighborhood of each object primitive. In our implementation, we determine the neighboring primitives via range query, and the radius is $0.05m$. 

The output of this module combines the label that indicates objectiveness, offsets in BB center, BB size, and BB orientation, and a semantic label. 


%% file: 07_training_details.tex
\subsection{Network Training}
\label{Section:Network:Training}

Training H3DNet employs a loss function with five objective terms:
\begin{align}
\min\limits_{\theta_g, \theta_p, \theta_c, \theta_o} &\quad  \lambda_g l_g(\theta_g) + \lambda_p l_p(\theta_g,\theta_p) + \lambda_{f} l_f(\theta_g, \theta_p, \theta_o) \nonumber \\
&\quad + \lambda_c l_c(\theta_g, \theta_p, \theta_c) + \lambda_o l_o(\theta_g, \theta_p, \theta_o)
\label{Eq:Total:Loss}
\end{align}
where $l_g$ trains the geometric primitive module $\theta_g$, $l_p$ trains the proposal module $\theta_p$, $l_f$ trains the potential function and refinement sub-network $\theta_o$, $l_c$ trains the classification sub-network $\theta_c$, and $l_o$ trains the refinement sub-network. The trade-off parameters $\lambda_g$, $\lambda_p$, $\lambda_f$, $\lambda_c$, and $\lambda_o$ are determined through 10-fold cross-validation. Intuitively, $l_c$, $l_o$ and $l_f$ provide end-to-end training of H3DNet, while $l_g$ and $l_p$ offer intermediate supervisions.

\noindent\textbf{Formulation.} Formulations of $l_g$, $l_p$, $l_c$, and $l_o$ follow common strategies in the literature. Specifically, both $l_g$ and $l_p$ utilize L2 regression losses and a cross-entropy loss for geometric primitive location and existence flag prediction, and initial proposal generation; $l_c$ applies a cross-entropy loss to train the object classification sub-network; $l_o$ employs L2 regression losses for predicting the shape offset, and a cross-entropy loss for predicting the semantic label. Since these four loss terms are quite standard, we leave the details to the supplemental material. 

$l_f$ seeks to match the local minimums of the potential function and the underlying ground-truth objects.  Specifically, consider a parametric potential function $f_{\Theta}(\bs{x})$ parameterized by $\Theta$. Consider a local minimum $\bs{x}_{\Theta}^{\star}$ which is a function of $\Theta$. Let $\bs{x}^{\gt}$ be the target location of $\bs{x}_{\Theta}^{\star}$. We define the following alignment potential to pull $\bs{x}_{\Theta}^{\star}$ to close to $\bs{x}^{\gt}$:
\begin{equation}
l_{m}(\bs{x}_{\Theta}^{\star},\bs{x}^{\gt}):= \|\bs{x}_{\Theta}^{\star}-\bs{x}^{\gt}\|^2.    
\end{equation}
The following proposition describes how to compute the derivatives of $l_m$ with respect to $\Theta$. The proof is deferred to the supp. material.
\begin{proposition}
The derivatives of $l_m$ with respect to $\Theta$ is given by
\begin{equation}
\frac{\partial l_m}{\partial \Theta}:= 2(\bs{x}_{\Theta}^{\star}-\bs{x}^{\gt})^T\cdot \frac{\partial \bs{x}_{\Theta}^{\star}}{\partial \Theta},\quad \frac{\partial \bs{x}_{\Theta}^{\star}}{\partial \Theta}:= -\Big(\frac{\partial^2 f_{\Theta}(\bs{x}^{\star})}{\partial^2 \bs{x}}\Big)^{-1}\cdot \frac{\partial^2 f_{\Theta}(\bs{x}^{\star})}{\partial \bs{x}\partial \Theta}.
\label{Eq:Derivatives}
\end{equation}
\label{Prop:IFM}
\end{proposition}
%
%

We proceed to use $l_m$ to define $l_f$. For each scene $S$, we denote the set of ground-truth objects and the set of local minimums of potential function $F_{S}$ as $\set{O}^{\gt}$ and $\set{O}^{\star}$, respectively. Note that $\set{O}^{\star}$ depends on the network parameters and hyper-parameters. Let $\set{C}_{S}\subset \set{O}^{\gt}\times \set{O}^{\star}$ collect the nearest object in $\set{O}^{\star}$ for each object in $\set{O}^{\gt}$. Consider a training set of scenes $\set{S}_{train}$, we define 
\begin{equation}
l_f:= \sum\limits_{S \in \set{S}_{train}}\sum\limits_{(o^{\star},o^{\gt})\in \set{C}_S} l_m(\bs{o}^{\star}, \bs{o}^{\gt}).  
\end{equation}
Computing the derivatives of $l_f$ with respect to the network parameters is a straightforward application of Prop.\ref{Prop:IFM}.

\noindent\textbf{Training.} We train H3DNet end-to-end and from scratch with the Adam optimizer~\cite{KingmaB14}. Please defer to the supplemental material for hyper-parameters used in training, such as learning rate etc.

%% file: 08_results.tex
\section{Experimental Results}
\label{Section:Results}

\begin{figure*}[t!]
\includegraphics[width=1.0\textwidth]{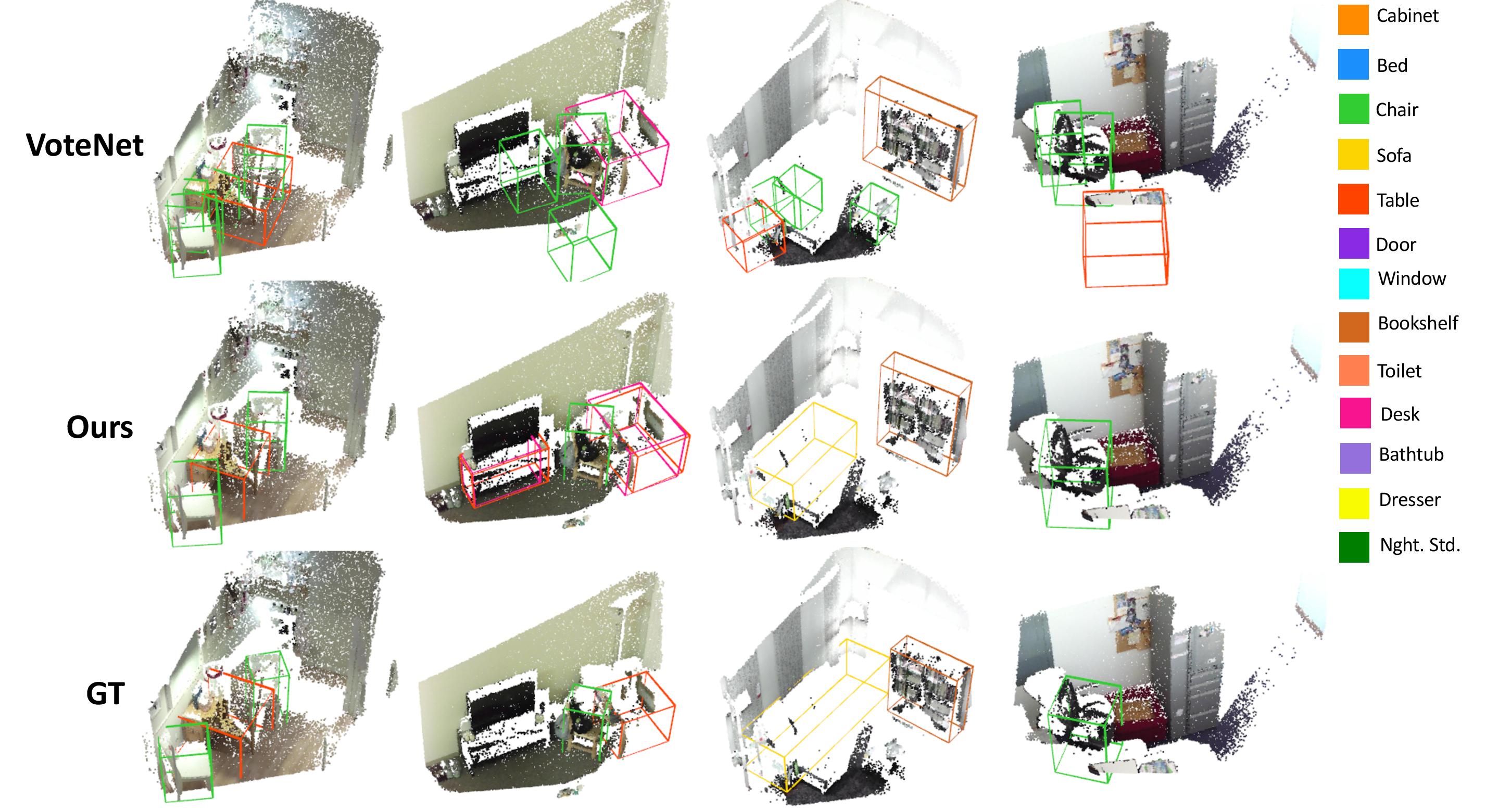}
\caption{Qualitative baseline comparisons on SUN RGB-D.}
\label{Figure:Qualitative:Result:Sun}
\end{figure*}

In this section, we first describe the experiment setup in Section~\ref{Subsection:Analysis}. Then, we compare our method with current state-of-the-art 3D object detection methods quantitatively, and analyze our results in Section~\ref{Subsection:Analysis}, where we show the importance of using geometric primitives and discuss our advantages. Finally, we show ablation results in Section~\ref{Subsection:Ablation:Study} and qualitative comparison in Figures~(\ref{Figure:Qualitative:Result:Sun}) and (\ref{Figure:Qualitative:Result:Scan}). More results and discussions can be found in the supplemental material.

\subsection{Experimental Setup}
\label{Subsection:Setup}

\noindent\textbf{Datasets.} We employ two popular datasets ScanNet V2\cite{Dai_2017_CVPR_scannet} and SUN RGB-D V1\cite{song2015sun}. 
ScanNet is a dataset of richly-annotated 3D reconstructions of indoor scenes. It contains 1513 indoor scenes annotated with per-point instance and semantic labels for 40 semantic classes. SUN RGB-D is a single-view RGB-D dataset for 3D scene understanding, which contains 10335 indoor RGB and depth images with per-point semantic labels and object bounding boxes. For both datasets, we use the same training/validation split and BB semantic classes (18 classes for ScanNet and 10 classes for SUN RGB-D) as in VoteNet\cite{qi2019votenet} and sub-sample 40000 points from every scene. 

\noindent\textbf{Evaluation protocol.} We use Average Precision(AP) and the mean of AP across all semantic classes (mAP)\cite{song2015sun}  under different IoU values (the minimum IoU to consider a positive match). Average precision computes the average precision value for recall value over 0 to 1. IoU is given by the ratio of the area of intersection and area of union of the predicted bounding box and ground truth bounding box. Specifically, we use AP/mAP@0.25 and AP/mAP@0.5. 


\begin{table*}[t!]
\caption{3D object detection results on ScanNet V2 val dataset. We show per-category results of average precision (AP) with 3D IoU threshold 0.25 as proposed by \cite{song2015sun}, and mean of AP across all semantic classes with 3D IoU threshold 0.25.} 
\begin{adjustbox}{width=\columnwidth,center}
 \begin{tabular}{l | c | c  c  c  c  c  c  c  c  c  c  c  c  c  c  c  c  c  c | c } 
 \hline
 & RGB & cab & bed & chair & sofa & tabl & door & wind & bkshf & pic & cntr & desk & curt & fridg & showr & toil & sink & bath & ofurn & mAP\\
 \hline
 3DSIS-5\cite{Hou_2019_CVPR_3D-SIS} & \cmark & 19.8 & 69.7 & 66.2 & 71.8 & 36.1 & 30.6 & 10.9 & 27.3 & 0.0 & 10.0 & 46.9 & 14.1 & 53.8 & 36.0 & 87.6 & 43.0 & 84.3 & 16.2 & 40.2 \\ 
3DSIS\cite{Hou_2019_CVPR_3D-SIS} & \xmark & 12.8 & 63.1 & 66.0 & 46.3 & 26.9 & 8.0 & 2.8 & 2.3 & 0.0 & 6.9 & 33.3 & 2.5 & 10.4 & 12.2 & 74.5 & 22.9 & 58.7 & 7.1 & 25.4  \\
Votenet\cite{qi2019votenet} &\xmark& 36.3 & 87.9 & 88.7 & 89.6 & 58.8 & 47.3 & 38.1 & 44.6 & 7.8 & 56.1 & 71.7 & 47.2 & 45.4 & 57.1 & 94.9 & 54.7 & 92.1 & 37.2 & 58.7 \\
\hline
 Ours & \xmark & \textbf{49.4} & \textbf{88.6} & \textbf{91.8} & \textbf{90.2} & \textbf{64.9} & \textbf{61.0} & \textbf{51.9} & \textbf{54.9} & \textbf{18.6} & \textbf{62.0} & \textbf{75.9} & \textbf{57.3} & \textbf{57.2} & \textbf{75.3} & \textbf{97.9} & \textbf{67.4} & \textbf{92.5} & \textbf{53.6} & \textbf{67.2} \\
 \hspace{0.1cm}\small{w\textbackslash o refine} & \xmark &  37.2 &
 89.3 & 88.4 & 88.5 & 64.4 & 53.0 & 44.2 & 42.2 & 11.1 & 51.2 & 59.8 &
 47.0 & 54.3 & 74.3 & 93.1 & 57.0 & 85.6 & 43.5 & 60.2 
 \\
 \hline
\end{tabular}
\label{Table:Quantitative:Result:ScanNetCat}
\end{adjustbox}
\end{table*}

\begin{figure*}[b!]
\includegraphics[width=1.0\textwidth]{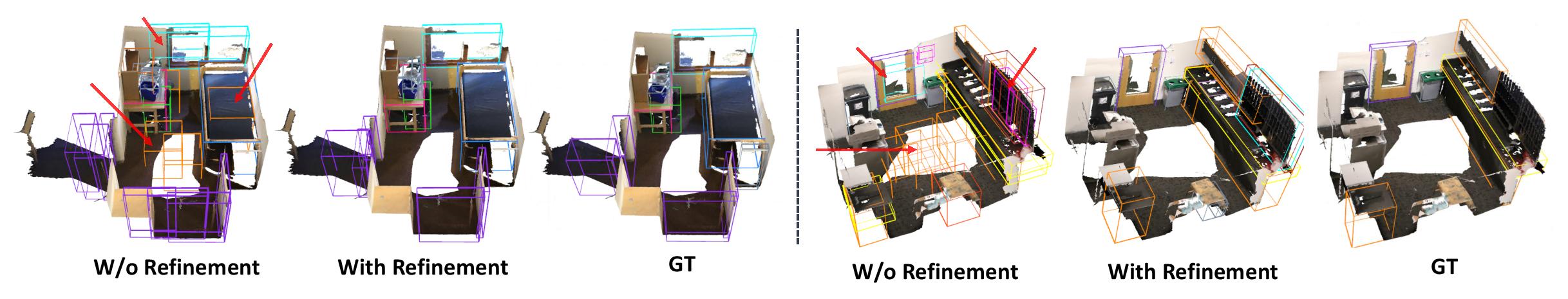}
\caption{\small{Effect of geometric primitive matching and refinement.}}
\label{Figure:Qual:Geometric:Constraints}
\end{figure*}

\noindent \textbf{Baseline Methods} We compare H3DNet with STAR approaches: VoteNet~\cite{qi2019votenet} is a geometric-only detector that combines deep point set networks and a voting procedure. GSPN\cite{yi2019gspn} uses a generative model for instance segmentation. Both 3D-SIS~\cite{Hou_2019_CVPR_3D-SIS} and DSS~\cite{Song_2016_CVPR} extract features from 2D images and 3D shapes to generate object proposals. F-PointNet~\cite{qi2018frustum} and 2D-Driven~\cite{Lahoud_2017_ICCV} first propose 2D detection regions and project them to 3D frustum for 3D detection. Cloud of gradient(COG)~\cite{Ren_2016_CVPR} integrates sliding windows with a 3D HoG-like feature. 





\begin{table*}[t!]
\caption{\textbf{Left:} 3D object detection results on ScanNetV2 val set. \textbf{Right:} results on SUN RGB-D V1 val set. We show mean of average precision (mAP) across all semantic classes with 3D IoU threshold 0.25 and 0.5. }
\begin{minipage}{0.49\linewidth}
\begin{adjustbox}{width=0.99\columnwidth, center}
\centering
 \begin{tabular}{l | c | c | c  } 
 \hline
 & Input  & mAP@0.25 & mAP@0.5  \\ 
 \hline
 DSS\cite{Song_2016_CVPR} & Geo $+$ RGB & 15.2 & 6.8 \\ 
 F-PointNet\cite{qi2018frustum} & Geo + RGB & 19.8 & 10.8 \\
 GSPN\cite{yi2019gspn} & Geo + RGB & 30.6 & 17.7\\
 3D-SIS \cite{Hou_2019_CVPR_3D-SIS} & Geo + 5 views & 40.2 & 22.5 \\
 VoteNet \cite{qi2019votenet} & Geo only & 58.7 & 33.5 \\
 \hline 
 Ours & Geo only & \textbf{67.2} & \textbf{48.1} \\ 
 \hspace{0.1cm}\small{w\textbackslash o refine} & Geo only & 60.2 & 37.3 \\ 
 \hline
\end{tabular}
\end{adjustbox}
\end{minipage}
\begin{minipage}{0.49\linewidth}
\begin{adjustbox}{width=0.94\columnwidth, center}
\centering
 \begin{tabular}{l | c | c | c  } 
 \hline
 & Input  & mAP@0.25 & mAP@0.5  \\
 \hline
 DSS\cite{Song_2016_CVPR} & Geo + RGB & 42.1 & - \\ 
 COG\cite{Ren_2016_CVPR} & Geo + RGB & 47.6 & -\\ 
 2D-driven\cite{Lahoud_2017_ICCV} & Geo + RGB &  45.1 & - \\
 F-PointNet\cite{qi2018frustum} & Geo + RGB & 54.0 & -\\
 VoteNet \cite{qi2019votenet} & Geo only & 57.7 &  32.9 \\
 \hline 
 Ours & Geo only & \textbf{60.1} & \textbf{39.0}\\
 \hspace{0.1cm}\small{w\textbackslash o refine} & Geo only & 58.5 & 34.2 \\
 \hline
\end{tabular}
\end{adjustbox}
\end{minipage}
\label{Table:Quantitative:Result:All}
\end{table*}

\subsection{Analysis of Results}
\label{Subsection:Analysis}

\begin{figure*}[b!]
\includegraphics[width=1.0\textwidth]{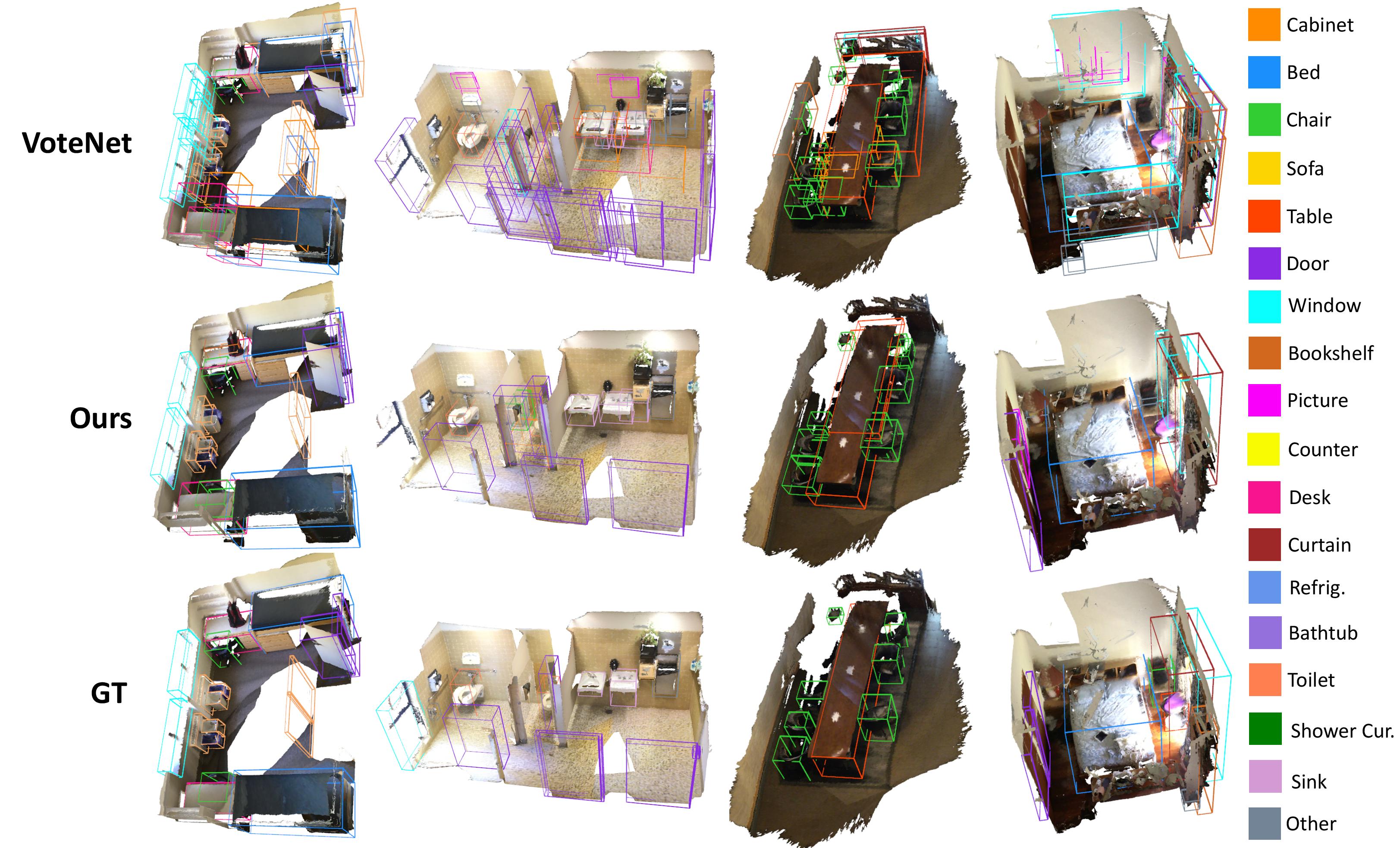}
\caption{Qualitative baseline comparisons on ScanNet V2.}
\label{Figure:Qualitative:Result:Scan}
\end{figure*}

As shown in Table \ref{Table:Quantitative:Result:All}, our approach leads to an average mAP score of 67.2\%, with 3D IoU threshold 0.25 (mAP@0.25), on ScanNet V2, which is 8.5\% better than the top-performing baseline approach~\cite{qi2019votenet}. In addition, our approach is 14.6\% better than the baseline approach~\cite{qi2019votenet} with 3D IoU threshold 0.5 (mAP@0.5). For SUN RGB-D, our approach gains 2.4\% and 6.1\% in terms of mAP, with 3D IoU threshold 0.25 and 0.5 respectively. 
On both datasets, the performance gains of our approach under mAP@0.5 are larger than those under mAP@0.25, meaning our approach offers more accurate predictions than baseline approaches. Such improvements are attributed to using an overcomplete set of geometric primitives and their associated features for generating and refining object proposals. We can also understand the relative less salient improvements on SUN RGB-D than ScanNet in a similar way, i.e., labels of the former are less accurate than the latter, and the strength of H3DNet is not fully utilized on SUN RGB-D. Except for the classification and refinement module, our approach shares similar computation pipeline and complexity with VoteNet. The computation on multiple descriptor towers and proposal modules can be paralleled, which should not increase computation overhead. In our implementation, our approach requires 0.058 seconds for the last module per scan. Conceptually, our approach requires 50\% more time compared to~\cite{qi2019votenet} but operates with a higher detection accuracy.

\noindent\textbf{Improvement on thin objects.} One limitation of the current top-performing baseline~\cite{qi2019votenet} is predicting thin objects in 3D scenes, such as doors, windows and pictures. In contrast, with face and edge primitives, H3DNet is able to extract better features for those thin objects. For example, the frames of window or picture provide dense edge feature, and physical texture of curtain or shower-curtain provide dense face/surface feature. As shown in Table \ref{Table:Quantitative:Result:ScanNetCat}, H3DNet leads to significant performance gains on thin objects, such as door (13.7\%), window (13.8\%), picture (10.8\%), curtain (10.1\%) and shower-curtain (18.2\%).

\noindent\textbf{Improvement on objects with dense geometric primitives.}
Across the individual object classes in ScanNet in Table \ref{Table:Quantitative:Result:ScanNetCat}, other than those thin objects, our approach also leads to significant performance gain on cabinet (13.1\%), table (6.1\%), bookshelf (10.3\%), refrigerator (11.8\%), sink (12.7\%) and other-furniture (16.4\%). One explanation is that the geometric shapes of these object classes possess rich planar structures and/or distinct edge structures, which contribute greatly on geometric primitive detection and object refinement.

\begin{table*}[b!]
\begin{minipage}{0.48\textwidth}
  \centering
  \includegraphics[width=0.8\textwidth]{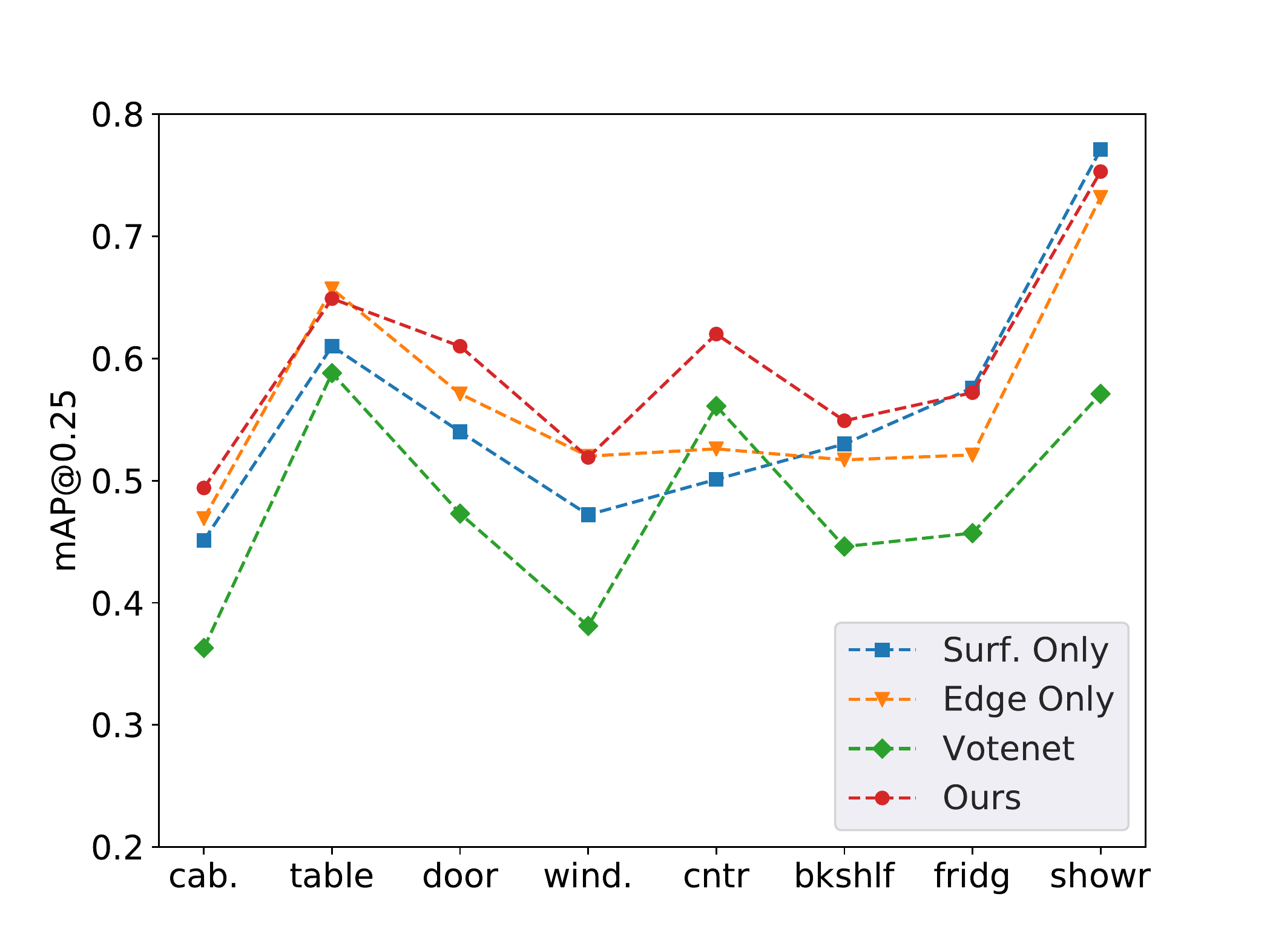}
  \captionof{figure}{Quantitative comparisons between VoteNet, our approach, ours with only face primitive and ours with only edge primitive, across sampled categories for ScanNet.}
  \label{Figure:Ablation:Primitive}
\end{minipage}
\hfill
\begin{minipage}{0.48\textwidth}
    \caption{Quantitative results without refining predicted center, size, semantic or object existence score for ScanNet, and without refining predicted angle for SUN RGB-D and differences compared with refining all.}
    \centering
  \begin{tabular}{l | c | c | c | c } 
 \hline
  & \multicolumn{2}{c}{mAP@0.25} & \multicolumn{2}{|c}{mAP@0.5} \\
 \hline
  w\textbackslash o center & 66.9 & -0.3 & 46.3 & -1.8\\ 
  w\textbackslash o size & 65.4 & -1.8 & 44.2 & -3.9\\ 
  w\textbackslash o semantic & 66.2 & -1.0 & 47.3 & -0.8\\ 
  w\textbackslash o existence & 65.2 & -1.8 & 45.1 & -3.0\\ 
  \hline\hline
  w\textbackslash o angle & 58.6 & -1.5 & 36.6 & -2.4 \\
  \hline
 \end{tabular}
  \label{Table:Ablation:Refinement}
\end{minipage}
\end{table*}

\noindent\textbf{Effect of primitive matching and refinement.} Using a distance function to refine object proposals and aggregating features of matching primitives are crucial for H3DNet. On ScanNet, merely classifying the initial proposals results in a 14.6\% drop on mAP 0.5. Figure~\ref{Figure:Qual:Geometric:Constraints} shows qualitative object detection results, which again justify the importance of optimizing and refining object proposals. 

\subsection{Ablation Study}
\label{Subsection:Ablation:Study}

\noindent\textbf{Effects of using different geometric primitives.} H3DNet can utilize different groups of geometric primitives for generating, classifying, and refining object proposals. Such choices have profound influences on the detected objects. As illustrated in Figure~\ref{Figure:Ablation:Primitive}, when only using BB edge primitives, we can see that objects with prominent edge features, i.e., window, possess accurate predictions. In contrast, objects with dense face/surface features, such as shower curtain, exhibit relative low prediction accuracy. However, these objects can be easily detected by activating BB face primitives. H3DNet, which combines BB centers, BB edge centers, and BB face centers, adds the strength of their generalization behaviors together. The resulting performance gains are salient when compared to using a single set of geometric primitives.  


\noindent\textbf{Effects of proposal refinement.} During object proposal refinement, object center, size, heading angle, semantic and existence are all optimized. As shown in Table \ref{Table:Ablation:Refinement}, without fine-tuning any of the geometric parameters of the detected objects, the performance drops, which shows the importance of this sub-module. 

\begin{table*}[t!]
\begin{minipage}{0.48\textwidth}
  \centering
  \includegraphics[width=0.8\textwidth]{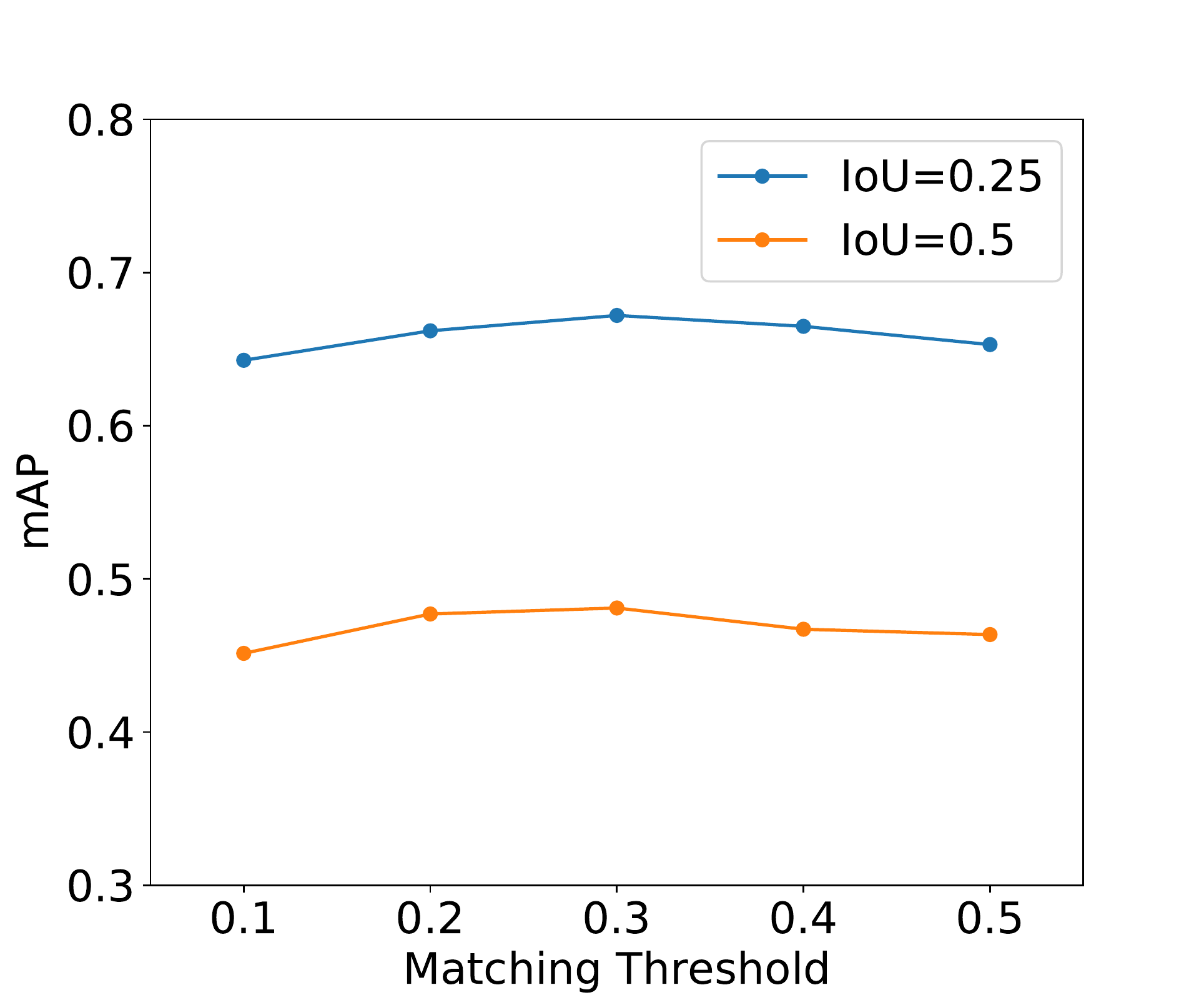}
  \captionof{figure}{Quantitative comparisons between different truncation threshold $\delta$ for ScanNet.}
  \label{Figure:Ablation:Threshold}
\end{minipage}
\hfill
\begin{minipage}{0.48\textwidth}
    \caption{Quantitative comparisons between different number of descriptor computation towers, among our approach and VoteNet, for ScanNet and SUN RGB-D.}
    \centering
  \begin{tabular}{c | c | c | c  } 
 \hline
 & \# of Towers  & mAP@0.25 & mAP@0.5 \\
 \hline
 \multirow{4}{2.2em}{Ours} 
  & 1 & 64.4 & 43.4 \\ 
  & 2 & 65.4 & 46.2 \\ 
  & 3 & 66.0 & 47.7 \\ 
  & 4 & 67.2 & 48.3 \\ 
 \hline
 \multirow{2}{2.2em}{Vote} & 4 (Scan) &  60.11 & 37.12 \\ 
 & 4 (SUN) & 57.5 & 32.1 \\
 \hline
 \end{tabular}
  \label{Table:Ablation:FeatureExtraction}
\end{minipage}
\end{table*}

\noindent\textbf{Effect of different truncation threshold} As shown in Figure \ref{Figure:Ablation:Threshold}, with different truncation values of $\delta$, results with mAP@0.25 and mAP@0.5 remain stable. It shows that our model is robust to different truncation threshold $\delta$. 

\noindent\textbf{Effect of multiple descriptor computation towers.} One hyper-parameter of H3DNet is the number of descriptor computation towers. Table \ref{Table:Ablation:FeatureExtraction} shows that adding more descriptor computation towers leads to better results, yet the performance gain of adding more descriptor computation towers quickly drops. Moreover, the performance gain of H3DNet from VoteNet comes from the hybrid set of geometric primitives and object proposal matching and refinement. For example, replacing the descriptor computation tower of VoteNet by the four descriptor computation towers of H3DNet only results in modest and no performance gains on ScanNet and SUN RGB-D, respectively (See Table \ref{Table:Ablation:FeatureExtraction}). 


%% file: 09_conclusions.tex
\section{Conclusions and Future Work}
\label{Section:Conclusions:Future:Work}
In this paper, we have introduced a novel 3D object detection approach that takes a 3D scene as input and outputs a collection of labeled and oriented bounding boxes. The key idea of our approach is to predict a hybrid and overcomplete set of geometric primitives and then fit the detected objects to these primitives and their associated features. Experimental results demonstrate the advantages of this approach on ScanNet and SUN RGB-D. In the future, we would like to apply this approach to other 3D scene understanding tasks such as instance segmentation and CAD model reconstruction. Another future direction is to integrate more geometric primitives, like BB corners, for 3D object detection. 

\noindent\textbf{Acknowledgement.} We would like to acknowledge the support from NSF DMS-1700234, a Gift from Snap Research, and a hardware donation from NVIDIA.

%% file: 100_supp_intro.tex
\section{Introduction}
\label{Section::Supp:intro}

This supplemental material provides the proof of Proposition 1 in Section \ref{Section::Supp:proof}, additional details on network architecture and loss functions in Section \ref{Section:Supp:Network},  more analysis and experiment results on geometric primitive prediction in Section \ref{Section:Supp:Primitive}, and more analysis and experiment results on 3D object detection in Section \ref{Section:Supp:Results}.

%% file: 10_supp_proof.tex
\section{Proof of Proposition 1}
\label{Section::Supp:proof}
We show the proof of Proposition 1 here.

With chain rule and equation (3) in the main paper, we can directly get:
\begin{equation}
    \frac{\partial l_m}{\partial \Theta}= 2(\bs{x}_{\Theta}^{\star}-\bs{x}^{\gt})^T\cdot \frac{\partial \bs{x}_{\Theta}^{\star}}{\partial \Theta} .
\end{equation}

Since $\bs{x}^{\star}$ is the local minimum of $f_{\Theta}(\bs{x})$, we have:
\begin{equation}
    \frac{\partial f_{\Theta}(\bs{x}^{\star})}{\partial \bs{x}} = 0 .
\end{equation}

Compute the derivatives of both sides w.r.t. $\Theta$, i.e. 
\begin{equation}
    \frac{\partial^2 f_{\Theta}(\bs{x}^{\star})}{\partial^2 \bs{x}} \cdot \frac{\partial \bs{x}^{\star}}{\partial \Theta} + \frac{\partial^2 f_{\Theta}(\bs{x}^{\star})}{\partial \bs{x}\partial \Theta} = 0 ,
\end{equation}
which leads to the equation (4) in the main paper.

%% file: 11_supp_network.tex
\section{Details on network architecture and loss functions}
\label{Section:Supp:Network}

\subsection{Network architecture details}
In the main paper, we mentioned that there are three modules in H3DNet: geometric primitive module, proposal generation module, and classification and refinement module. We will discuss each module in detail.

\begin{figure}[t!]
\centering
\includegraphics[width=0.99\textwidth]{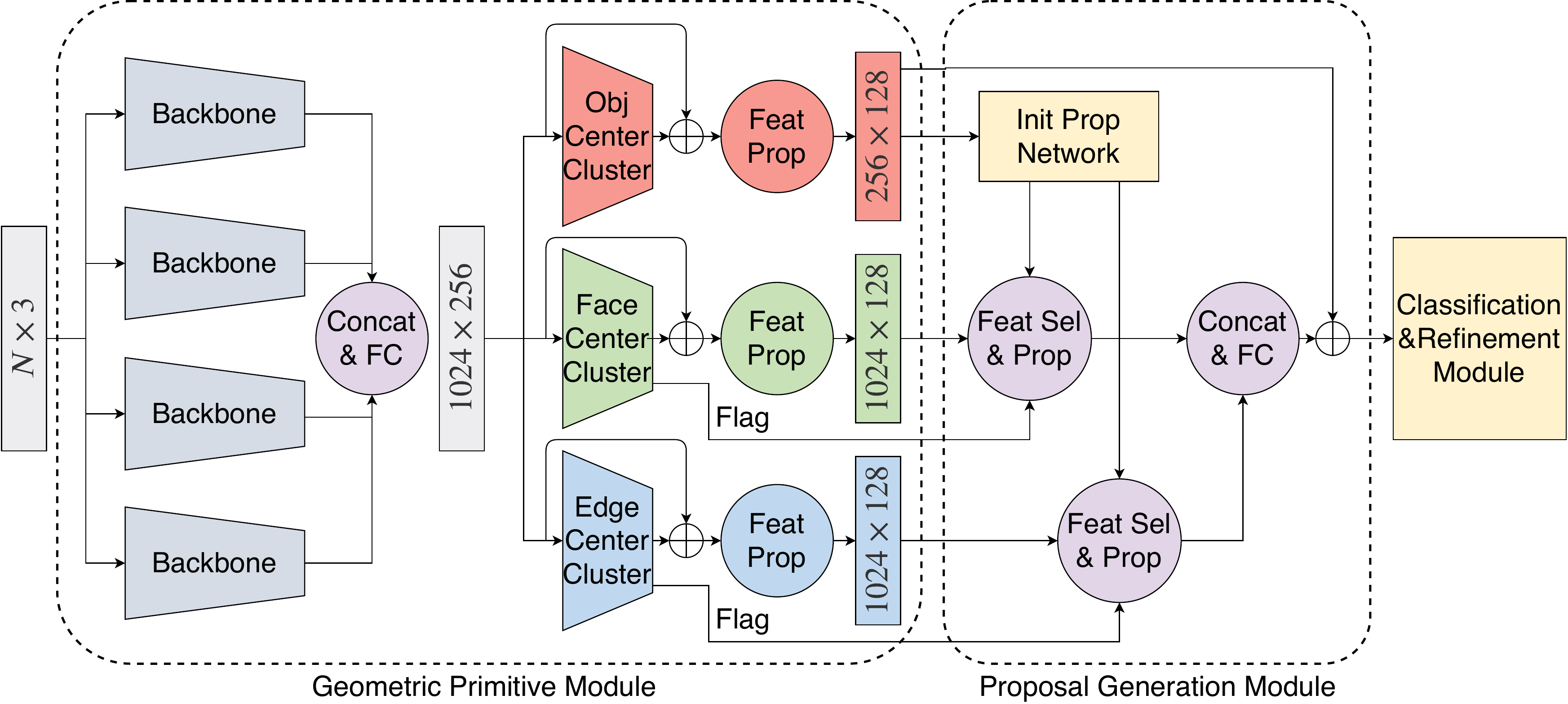}
\caption{\small{The pipeline of H3DNet. N represents the number of points of input point clouds, and we use 40000 for both datasets.}}
\label{Figure:Pipeline}
\end{figure}

The geometric primitive module first uses a tower of multiple backbone networks to extract down-sampled per-point feature, as shown in Figure \ref{Figure:Pipeline}. The backbone network, which is based on PointNet++~\cite{qi2017pointnet++}, was borrowed from~\cite{qi2019votenet}, and the same network configurations are used in our implementation. For all four backbone networks, we use the same index to sample 1024 points from input point clouds with 40000 points. We then concatenate the features for each point and then use two fully connected layers to reduce the feature dimension to 256. This hybrid feature then feeds into a cluster network, which contains three fully connected layers to predict an offset vector between each point and its corresponding center, i.e., object center, face center, and edge center. For face and edge primitive, we also predict a flag that indicates whether a point is close to a primitive or not. The cluster network also produces a residual feature vector, which will be added to the input feature vector. Finally, we use a set abstraction layer~\cite{qi2017pointnet++}, followed by four layers of multilayer perceptron (MLP) after the max-pooling in each local region to propagate features. For object centers, we sub-sampled 256 points for initial proposal generation, using furthest-point-sampling. For face and edge centers, we use the propagated features to predict a point-wise offset vector to refine the center prediction, and a point-wise semantic label to add semantic information in features of geometric primitives.

As shown in Figure \ref{Figure:Pipeline}, we then use three layers of MLP to generate initial object proposals. We use the same configuration as in~\cite{qi2019votenet}. As mentioned in the main paper, we then associate each initial object proposal with an overcomplete set of geometric
primitives based on the local minimums of the distance function. However, the detected geometric primitives are firstly selected with the predicted flag, which indicates whether a point is close to a primitive or not. Again, we use a set abstraction layer~\cite{qi2017pointnet++}, followed by four layers of multilayer perceptron (MLP) after the max-pooling in each local region (i.e., a query ball with radius 0.5m), to propagate features between the predicted geometric primitives and the corresponding primitives of an object proposal. The propagated features are then concatenated and fed into a two-layer MLP for the final object proposals.

The last module is the classification and refinement module. It contains three layers of MLP. We add the feature generated from the proposal module with the object center feature generated in the primitive module, and feed it to the last module to acquire the final object proposal, including an object indicator, offset vectors to refine the BB center, BB size, and BB orientation, and a semantic label. The object indicator is used to determine whether an object exists in the scene or not. 

\subsection{Loss function details}
As mentioned in the main paper, the network is trained end-to-end with a multi-task loss function with five major objective terms. We will discuss each objective term in detail.

\begin{align}
\centering
l_g = l_{vote} + \lambda_1 l_{flag} + \lambda_2 l_{res} + \lambda_3 l_{sem}
\label{Eq:lgloss}
\end{align}

$l_g$ trains the geometric primitive module. Each primitive has its own objective. For object center offset, face center offset and edge center offset prediction, we adopt the same voting loss defined in~\cite{qi2019votenet}. As shown in equation \ref{Eq:lgloss}, for face and edge center, we add $l_{flag}$ for flag prediction, $l_{res}$ for point-wise center offset prediction (i.e. used in center refinement), and $l_{sem}$ for point-wise center semantic label prediction. We use a L1 loss, defined in~\cite{qi2019votenet}, for $l_{res}$, and a standard cross-entropy loss for $l_{flag}$ and $l_{sem}$. For equation \ref{Eq:lgloss}, we weight the losses so that they are in similar scales with $\lambda_1 = 3$, $\lambda_2 = 0.1$, and $\lambda_3 = 0.1$.

$l_p$ trains the proposal module $\theta_p$, which contains an objectness loss, a 3D bounding box estimation loss, and a semantic classification loss for initial object proposal generation. We adopt the same loss function defined in~\cite{qi2019votenet}. $l_f$ is the distance function defined in the main paper. $l_c$ is a standard cross-entropy loss, which trains the classification sub-network, and $l_o$ contains a cross-entropy loss for semantic label prediction, and an L2 regression loss for BB center, BB size, and BB orientation offset vector prediction. In our experiments, we set the trade-off parameters mentioned in the main paper, $\lambda_g$, $\lambda_p$, $\lambda_f$, $\lambda_c$, and $\lambda_o$ to 1.

\subsection{Training details}
Our network is implemented in PyTorch and optimized using Adam. The batch size is 8 and the number of epochs is 360. For ScanNet, the learning rate is initialized with 1e-2 and decreases by 10 times after 80, 140, 200, 240 epochs respectively. The learning rate of SUN RGB-D starts with 1e-3 and decreases by 10 times after 160, 220, 260 epochs respectively.

%% file: 12_supp_primitive.tex
\section{Geometric primitive prediction results and analysis}
\label{Section:Supp:Primitive}

\begin{figure}[b!]
\centering
\includegraphics[width=1.0\textwidth]{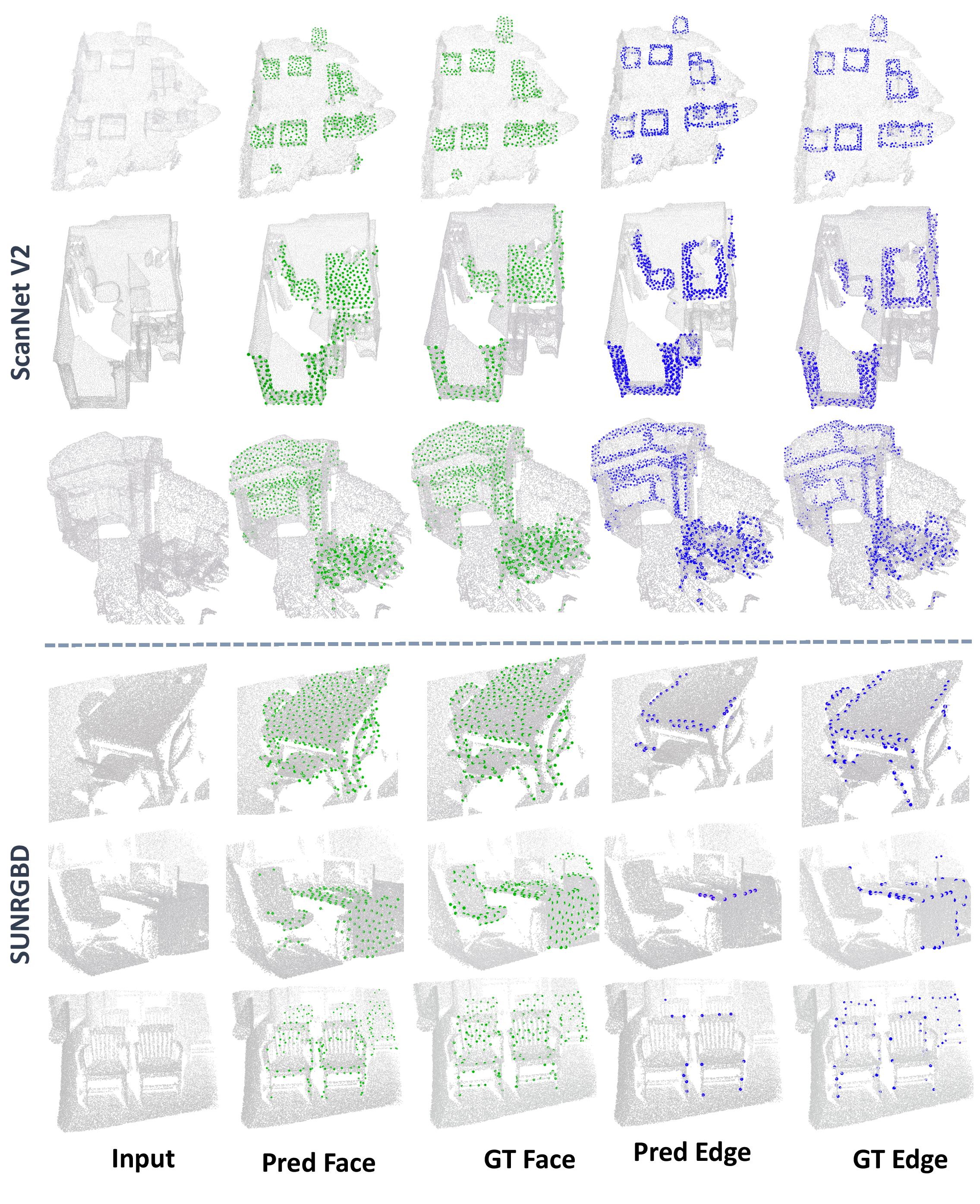}
\caption{\small{Qualitative examples for detected geometric primitives (face, edge).}}
\label{Figure:supp:cue}
\end{figure}

\subsection{Dataset Statistics}

\begin{table*}[b!]
\caption{Average number of edges and faces labelled per object in the ScanNet training dataset for different categories.} 
\begin{adjustbox}{width=\columnwidth,center}
 \begin{tabular}{| c | c  c  c  c  c  c  c  c  c  c  c  c  c  c  c  c  c  c | c |} 
 \hline
 type & cab & bed & chair & sofa & tabl & door & wind & bkshf & pic & cntr & desk & curt & fridg & showr & toil & sink & bath & ofurn & Avg \\
 \hline
 Face & 1.96 & 3.99 & 1.74 & 3.69 & 2.61 & 1.71 & 1.84 & 3.02 & 0.75 & 2.84 & 2.85 & 1.97 & 2.23 & 2.04 & 2.27 & 1.24 & 4.30 & 1.63 & 2.37\\ 
Edge & 5.60 & 7.10 & 6.33 & 7.48 & 7.64 & 3.76 & 5.19 & 6.89 & 2.80 & 6.90 & 7.62 & 5.40 & 6.17 & 4.62 & 7.95 & 6.23 & 9.72 & 5.03 & 6.25\\
 \hline
\end{tabular}
\end{adjustbox}
\label{Table:Quantitative:Result:ScanNetTrain}
\end{table*}

\begin{table*}[t!]
\caption{Average number of edges and faces labelled per object in the SUN RGB-D training dataset for different categories.} 
\begin{adjustbox}{width=0.8\columnwidth,center}
 \begin{tabular}{| c | c  c  c  c  c  c  c  c  c  c | c |} 
 \hline
  type & bathtub & bed & bkshf & chair & desk & drser & nigtstd &sofa &table & toilet & Avg \\
 \hline
 Face & 4.42 & 4.22 & 2.21 & 3.61 & 3.64 & 2.89 & 1.04 & 4.27 & 3.57 & 4.21 & 3.41 \\ 
Edge &  3.04 & 2.07 & 1.50 & 1.80 & 3.12 & 2.44 & 1.16 & 1.93 & 2.83 & 1.91 & 2.18 \\
 \hline
\end{tabular}
\end{adjustbox}
\label{Table:Quantitative:Result:SUNTrain}
\end{table*}

For an object with a 3D bounding box label, its maximum number of boundary faces is 6, and the maximum number of boundary edges is 12. In a real 3D scan, some faces or edges are not visible due to occlusion or irregular-shaped objects. As shown in Table \ref{Table:Quantitative:Result:ScanNetTrain} and \ref{Table:Quantitative:Result:SUNTrain}, we can see that on both datasets, there are dense labels for faces and edges. However, we can see that the edge labels per object in SUN RGB-D is significantly fewer than in ScanNet. Based on our observation, labels of the 3D bounding boxes in SUN RGB-D are less accurate than in ScanNet, and without per point instance labels, it is much more difficult to generate accurate and dense labels in SUN RGB-D.

\subsection{Qualitative Results}
We show some qualitative examples for detected geometric primitives in Figure \ref{Figure:supp:cue}. For better visualization purposes, we highlight the detected points if the predicted flag is valid. Most of the examples show that our model performs reasonably well on geometric primitive detection. However, the predictions on edges in SUN RGB-D are sparse due to the lack of labels in training data.

\begin{table*}[b!]
\begin{minipage}{0.48\textwidth}
  \centering
  \includegraphics[width=\textwidth]{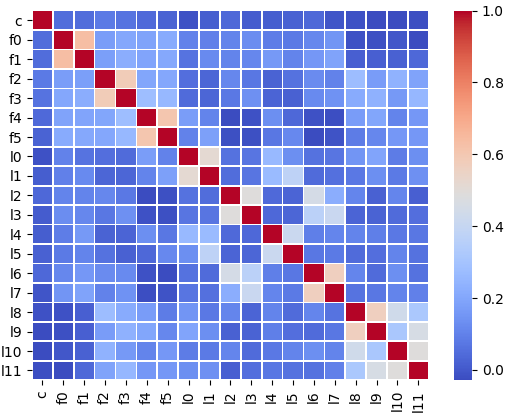}
\end{minipage}
\hfill
\begin{minipage}{0.48\textwidth}
  \centering
  \includegraphics[width=\textwidth]{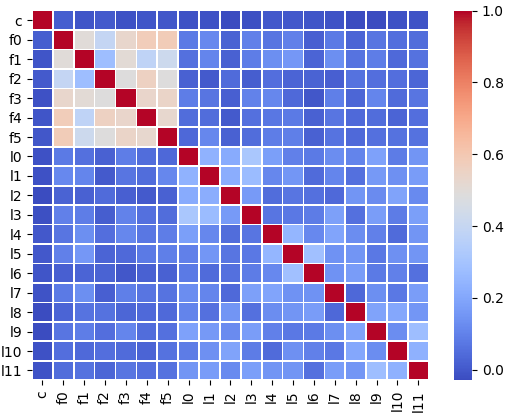}
\end{minipage}
 \captionof{figure}{\textbf{Left:} covariance matrix of ScanNet. \textbf{Right:} covariance matrix of SUN RGB-D. c represents object center, f0-f6 represent 6 BB face centers, and l0-l11 represent 12 BB edge centers.}
 \label{Figure:Quantitative:cov}
\end{table*}

\begin{table*}[t!]
\caption{Prediction accuracy of the location of geometric primitives, i.e. face and edge centers, across different categories for ScanNet. For each target primitive, if there is a prediction within 0.3m, we count it as a correct prediction.} 
\begin{adjustbox}{width=\columnwidth,center}
 \begin{tabular}{| c | c  c  c  c  c  c  c  c  c  c  c  c  c  c  c  c  c  c | c |} 
 \hline
 type & cab & bed & chair & sofa & tabl & door & wind & bkshf & pic & cntr & desk & curt & fridg & showr & toil & sink & bath & ofurn\\
 \hline
 Face & 0.89 & 0.86 & 0.97 & 0.91 & 0.94 & 0.93 & 0.83 & 0.86 & 0.53 & 0.93 & 0.88 & 0.91 & 0.99 & 0.98 & 0.98 & 0.98 & 0.93 & 0.84\\ 
Edge & 0.92 & 0.88 & 1.00 & 0.98 & 1.00 & 0.91 & 0.79 & 0.88 & 0.81 & 0.95 & 0.96 & 0.81 & 0.96 & 0.97 & 1.00 & 1.00 & 0.99 & 0.76\\
 \hline
\end{tabular}
\end{adjustbox}
\label{Table:Quantitative:Result:ScanNetPrimi}
\end{table*}

\begin{table*}[t]
\caption{Prediction accuracy of the location of geometric primitives, i.e. face and edge centers, across different categories for SUN RGB-D. For each target primitive, if there is a prediction within 0.3m, we count it as a correct prediction.} 
\begin{adjustbox}{width=0.8\columnwidth,center}
 \begin{tabular}{| c | c  c  c  c  c  c  c  c  c  c | c |} 
 \hline
 type & bathtub & bed & bkshf & chair & desk & drser & nigtstd &sofa &table & toilet \\
 \hline
 Face & 0.72 & 0.57 & 0.11 & 0.83 & 0.57 & 0.29 & 0.15 & 0.53 & 0.68 & 0.91 \\ 
Edge & 0.21 & 0.04 & 0.02 & 0.18 & 0.29 & 0.10 & 0.00 & 0.04 & 0.45 & 0.12 \\
 \hline
\end{tabular}
\end{adjustbox}
\label{Table:Quantitative:Result:SUNPrimi}
\end{table*}

\begin{table*}
\caption{3D object detection results on SUN RGB-D val dataset. We show per-category results of average precision (AP) with 3D IoU threshold 0.25 as proposed by \cite{song2015sun}, and mean of AP across all semantic classes.  Note that both COG \cite{Ren_2016_CVPR} and 2D-driven \cite{Lahoud_2017_ICCV} use room layout context to boost performance. For fair comparison with previous methods, the evaluation is on the SUN RGB-D V1 data.}
\begin{adjustbox}{width=\columnwidth,center}
 \begin{tabular}{l | c | c  c  c  c  c  c  c  c  c  c | c | c  } 
 \hline
 & RGB & bathtub & bed & bkshf & chair & desk & drser & nigtstd &sofa &table & toilet & mAP.25 \\
 \hline
 DSS\cite{Song_2016_CVPR} & \cmark& 44.2 & 78.8 & 11.9 & 61.2 & 20.5 & 6.4& 15.4& 53.5& 50.3& 78.9 & 42.1 \\ 
 COG\cite{Ren_2016_CVPR} & \cmark & 58.3 & 63.7 & 31.8 & 62.2 & 45.2 & 15.5 & 27.4& 51.0 & 51.3 & 70.1 & 47.6 \\ 
 2D-driven\cite{Lahoud_2017_ICCV} & \cmark & 43.5 & 64.5 & 31.4 & 48.3 & 27.9 & 25.9 & 41.9 & 50.4 & 37.0 & 80.4 & 45.1 \\
 F-PointNet\cite{qi2018frustum} & \cmark & 43.3 & 81.1 & 33.3 & 64.2 & 24.7 & 32.0 & 58.1 & 61.1 & 51.1 & 90.9 & 54.0 \\
 VoteNet \cite{qi2019votenet} &\xmark & \textbf{74.7} & 83.0 & 28.8 & 75.3 & 22.0 & 29.8 & 62.2 & 64.0 & 47.3 & 90.1 & 57.7  \\
 \hline 
 Ours & \xmark & 73.8 & 85.6 & 31.0 & \textbf{76.7} & \textbf{29.6} & \textbf{33.4} & \textbf{65.5} & \textbf{66.5} & 50.8 & 88.2 & \textbf{60.1} \\
 \hspace{0.1cm}\small{w\textbackslash o refine} & \xmark & 74.1 & \textbf{86.4} & \textbf{31.3} & 76.1 & 27.1 & 26.3 & 57.9 & 64.9 & \textbf{51.6} & \textbf{89.3} & 58.5 \\
 \hline
\end{tabular}
\end{adjustbox}
\label{Table:Quantitative:Result:SUNCAT}
\end{table*}

\begin{table*}
\caption{3D object detection results on SUN RGB-D val dataset. We show per-category results of average precision (AP) with 3D IoU threshold 0.5 as proposed by \cite{song2015sun}, and mean of AP. The evaluation is on the SUN RGB-D V1 data.}
\begin{adjustbox}{width=\columnwidth,center}
 \begin{tabular}{l | c | c  c  c  c  c  c  c  c  c  c | c | c  } 
 \hline
 & RGB & bathtub & bed & bkshf & chair & desk & drser & nigtstd &sofa &table & toilet & mAP.25 \\
 \hline
 VoteNet \cite{qi2019votenet} &\xmark & \textbf{49.9} & 47.3 & 4.6 & 54.1 & 5.2 & 13.6 & 35.0 & 41.4 & 19.7 & 58.6 & 32.9   \\
 \hline 
 Ours & \xmark & 47.6 & \textbf{52.9} & \textbf{8.6} & \textbf{60.1} & \textbf{8.4} & \textbf{20.6} & \textbf{45.6} & \textbf{50.4} & \textbf{27.1} & \textbf{69.1} & \textbf{39.0} \\
 \hspace{0.1cm}\small{w\textbackslash o refine} & \xmark & 48.9 & 50.6 & 5.0 & 55.6 & 6.3 & 14.6 & 32.7 & 45.1 & 23.3 & 60.1 & 34.2 \\
 \hline
\end{tabular}
\end{adjustbox}
\label{Table:Quantitative:Result:SUN05}
\end{table*}

\begin{table*}
\caption{3D object detection results on ScanNet V2 val dataset. We show per-category results of average precision (AP) with 3D IoU threshold 0.5 as proposed by \cite{song2015sun}, and mean of AP across all semantic classes with 3D IoU threshold 0.5.} 
\begin{adjustbox}{width=\columnwidth,center}
 \begin{tabular}{l | c | c  c  c  c  c  c  c  c  c  c  c  c  c  c  c  c  c  c | c } 
 \hline
 & RGB & cab & bed & chair & sofa & tabl & door & wind & bkshf & pic & cntr & desk & curt & fridg & showr & toil & sink & bath & ofurn & mAP\\
 \hline
 3DSIS-5\cite{Hou_2019_CVPR_3D-SIS} & \cmark & 5.73 & 50.28 & 52.59 & 55.43 & 21.96 & 10.88 & 0.00 & 13.18 & 0.00 & 0.00 & 23.62 & 2.61 & 24.54 & 0.82 & 71.79 & 8.94 & 56.40 & 6.87 & 22.53 \\ 
3DSIS\cite{Hou_2019_CVPR_3D-SIS} & \xmark & 5.06 & 42.19 & 50.11 & 31.75 & 15.12 & 1.38 & 0.00 & 1.44 & 0.00 & 0.00 & 13.66 & 0.00 & 2.63 & 3.00 & 56.75 & 8.68 & 28.52 & 2.55 & 14.60 \\
Votenet\cite{qi2019votenet} &\xmark& 8.1 & 76.1 & 67.2 & 68.8 & 42.4 & 15.3 & 6.4 & 28.0 & 1.3 & 9.5 & 37.5 & 11.6 & 27.8 & 10.0 & 86.5 & 16.8 & 78.9 & 11.7 & 33.5 \\
\hline
 Ours & \xmark & \textbf{20.5} & \textbf{79.7} & \textbf{80.1} & \textbf{79.6} & \textbf{56.2} & \textbf{29.0} & \textbf{21.3} & \textbf{45.5} & \textbf{4.2} & \textbf{33.5} & \textbf{50.6} & \textbf{37.3} & \textbf{41.4} & \textbf{37.0} & \textbf{89.1} & \textbf{35.1} & \textbf{90.2} & \textbf{35.4} & \textbf{48.1} \\
 \hspace{0.1cm}\small{w\textbackslash o refine} & \xmark & 12.4 & 80.8 & 69.3 & 71.8 & 42.6 & 19.5 & 12.3 & 26.1 &2.4 & 15.7 & 27.3 & 32.6 & 29.5 & 33.6 & 79.1 & 23.0 & 74.0 & 18.9 & 37.3\\
 \hline
\end{tabular}
\end{adjustbox}
\label{Table:Quantitative:Result:ScanNet05}
\end{table*}

\subsection{Quantitative Analysis}

In this section, we provide an empirical analysis of the benefits of different geometric primitives. The primary observations are:
\begin{itemize}
\item different geometric primitives are suitable for various object categories;
\item the bias of the predictions are generally smaller than the variance of the predictions;
\item errors in different predictions are mostly uncorrelated. 
\end{itemize}
When aggregating different predictions together, the truncated L2 loss function can prune outlier predictions. Therefore, we obtain a variance reduction and improved prediction accuracy.   

\noindent\textbf{Prediction errors under different geometric primtiives.} Prediction accuracy of geometric primitives, i.e. face and edge centers, is shown in Table \ref{Table:Quantitative:Result:ScanNetPrimi} and \ref{Table:Quantitative:Result:SUNPrimi}. Since we are predicting an overcomplete set of geometric primitives, we only show the prediction accuracy of detected geometric primitives near the target ground-truth primitives. As shown in Table \ref{Table:Quantitative:Result:ScanNetPrimi}, for most categories, the prediction accuracy of edge center primitives is higher. However, for some categories, like window and curtain, we observe higher accuracy with face center primitive. It shows the different error distributions of BB face centers and BB edge centers, which demonstrate the importance of utilizing multiple geometric primitives. The prediction accuracy of geometric primitives in SUN RGB-D is significantly lower than in ScanNet, especially for edge center labels. This is again caused by sparse and inaccurate labels in training data.

\begin{figure}
\includegraphics[width=1.0\textwidth]{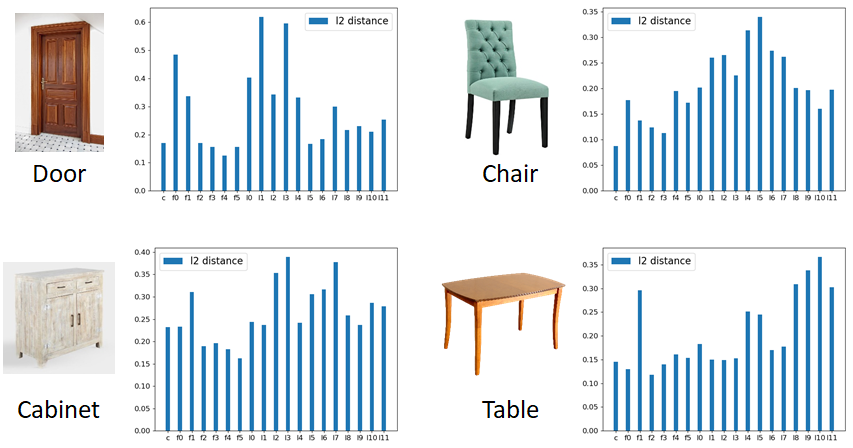}
\caption{Prediction errors of different geometric primitives under four different categories of the ScanNet v2 dataset.}
\label{Figure:Prediction:Error:ScanNet}
\end{figure}

Figure~\ref{Figure:Prediction:Error:ScanNet} shows the prediction errors of different geometric primitives under four categories of the test set of ScanNet v2 dataset. We can see that the error patterns are different when varying the categories. This again shows the benefits of having different types of geometric primitives as intermediate supervision. Note that each prediction error is a 3D vector, and we report its norm as the error.  

\noindent\textbf{Bias is smaller than the variance.} Figure shows that bias and variance of each geometric primitive with respect to the test sets of ScanNet v2 and SUNRBG-D. Here we report the norm of the expecation and the spectrum norm of each 3x3 co-variance matrix. We can see that generally the bias is smaller than the variance. Moreover, this ratio is even smaller on face centers and edge centers than the box center. This shows the usefulness of hybrid geometric primitives. 

The reason why bias is smaller than the variance can be understood from the perspective that during training, the training error is generally smaller than the testing error. 

\noindent\textbf{Different predictions are mostly uncorrelated.} In Figure \ref{Figure:Quantitative:cov}, we visualize the covariance matrix of the error distribution for 19 geometric primitives. For each target geometric primitive, we measure the Euclidean distance to the nearest predicted point and concatenate the results across every object in every testing scene. As shown in Figure \ref{Figure:Quantitative:cov}, the error distributions across all 19 geometric primitives are uncorrelated in ScanNet. Although the error distributions of 6 BB face centers in SUN RGB-D are slightly correlated, other geometric primitives are still uncorrelated. 

One interpretation is that in the over-parameterized regime, the optimized network weights are close to the initial network weights. Therefore, if the initial weights are independent, then the optimized weights are also approximately independent. It follows that different predictions are not strongly correlated. We leave a detailed theoretical analysis for future work.

\begin{figure}
\centering
\includegraphics[width=0.49\textwidth]{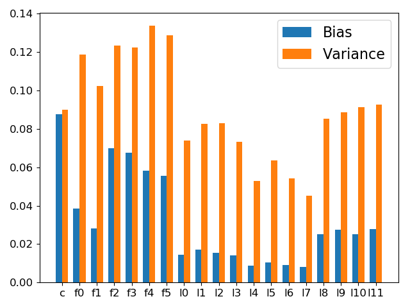}
\includegraphics[width=0.49\textwidth]{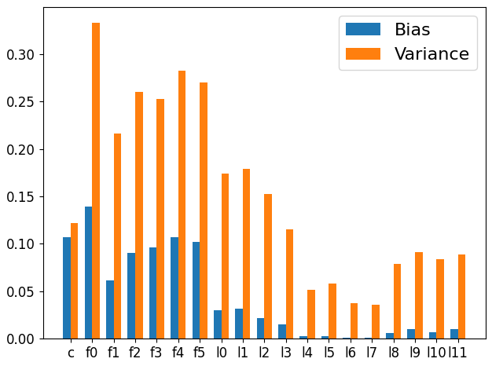}
\caption{(Left): magnitudes of bias and variance (square-root) of geometric primitive predictions on ScanNet. (Right): magnitudes of bias and variance (square-root) of geometric primitive predictions on SUNRBGD
}
\label{Figure:BV:Tradeoff}
\end{figure}

\noindent\textbf{Variance reduction and improved accuracy.}
For simplicity, we focus on analyzing the error of the predicted box center. The analysis of both box parameters are similar. For box center, the prediction is simply a weighted average of all predictions:
\begin{equation}
\bs{x}_{pred} = \frac{\bs{y}_{\cbox}+\beta_{\face}\sum\limits_{i=1}^6 \bs{y}_{\face,i}+\beta_{\edge}\sum\limits_{i=1}^{12}\bs{y}_{\edge,i}}{1 + 6\beta_{\face} + 12 \beta_{\edge}}    
\end{equation}
where $\bs{y}_{\cbox}$ denotes the box center prediction; $\bs{y}_{\face,i}$ denotes the prediction of the $i$-th face center; $\bs{y}_{\edge,i}$ denotes the prediction of the $i$-th edge center. Denote the norm of the bias vector of $\bs{x}_{pred}$ as $b_{pred}$. It is clear that
$$
b_{pred} \leq \frac{b_{\cbox} + \beta_{\face}\sum\limits_{i=1}^6 b_{\face,i} + \beta_{\edge}\sum\limits_{i=1}^{12}b_{\edge,i}}{1 + 6\beta_{\face} + 12 \beta_{\edge}}
$$
where $b_{\face,i}$ and $b_{\edge,i}$ are the norms of the bias vectors of $\bs{y}_{\face,i}$ and $\bs{y}_{\edge,i}$, respectively. It is clear that $b_{pred}$ is smaller than the largest bias of each individual prediction. 

The variance of $\bs{x}_{pred}$ is given by
$$
V[\bs{x}_{pred}] = \frac{V[\bs{y}_{\cbox}] + \beta_{\face}\sum\limits_{i=1}^{6}V[\bs{y}_{\face,i}] + \beta_{\edge}\sum\limits_{i=1}^{12}V[\bs{y}_{\edge, i}]}{(1 + 6\beta_{\face} + 12 \beta_{\edge})^2}
$$
Therefore, with suitable chosen trade-off parameters, we obtain a reduction in variance. Combing the fact that the bias is smaller than the variance, $x_{pred}$ is expected to lead to improved accuracy.  

%% file: 13_supp_result.tex
\section{More Analysis Experiments}
\label{Section:Supp:Results}
\subsection{More quantitative results}
We show the per-category results on ScanNet with 3D IoU threshold 0.5 in Table \ref{Table:Quantitative:Result:ScanNet05}, and the per-category results on SUN RGB-D with both 3D IoU threshold 0.25 and 0.5 in Table \ref{Table:Quantitative:Result:SUNCAT} and \ref{Table:Quantitative:Result:SUN05}. For accurate object detection, our approach outperforms the baseline approaches significantly. For thin objects in ScanNet, our approach can gain 14.9\%, 24.0\%, 25.7\%, and 27.0\% increase on Window, Counter, Curtain, and Shower-curtain. Again, these improvements are achieved by using an overcomplete set of hybrid geometric primitives and their associated features for generating and refining object proposals. Such performance gain can also be observed for SUN RGB-D in Table \ref{Table:Quantitative:Result:SUN05}, where our approach performs significantly better on more accurate object detection.

\subsection{More qualitative results}

\begin{figure}
\includegraphics[width=1.0\textwidth]{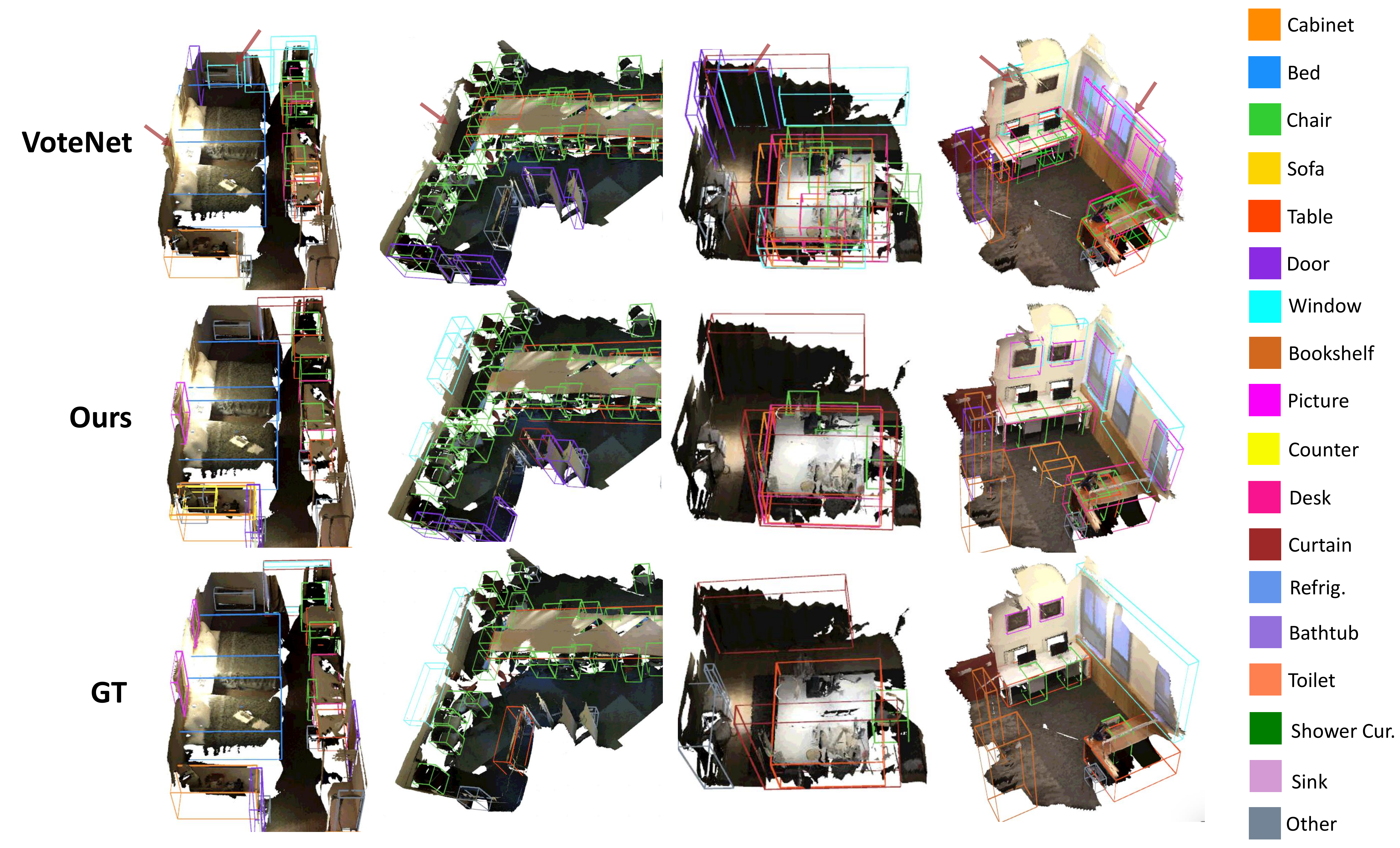}
\caption{\small{Qualitative evaluation on thin object detection. Red arrows are used to highlight the thin objects.}}
\label{Figure:Qual:thin}
\end{figure}

\begin{figure}[b!]
\includegraphics[width=1.0\textwidth]{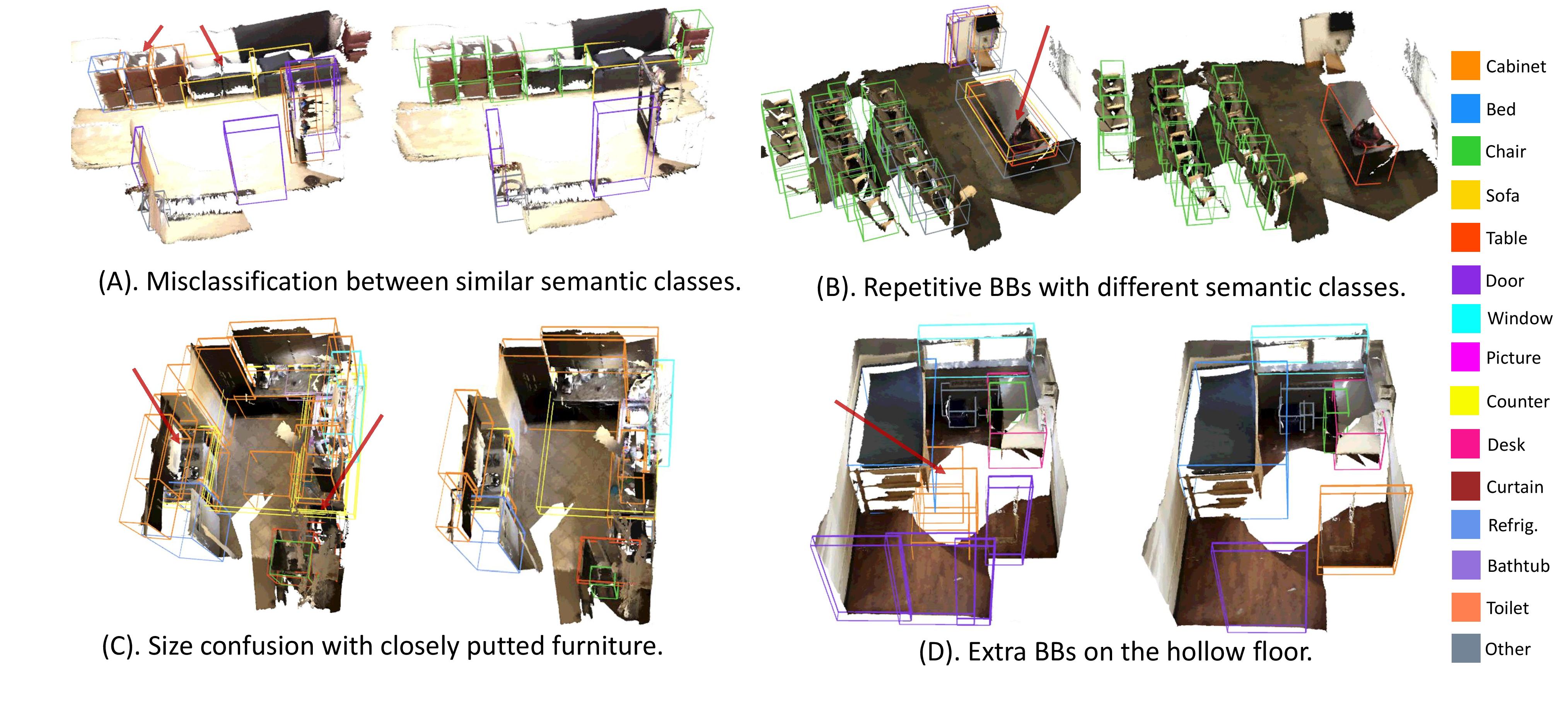}
\caption{\small{Samples of failure cases.}}
\label{Figure:Qual:failures}
\end{figure}

\begin{figure}
\centering

\includegraphics[width=0.9\textwidth]{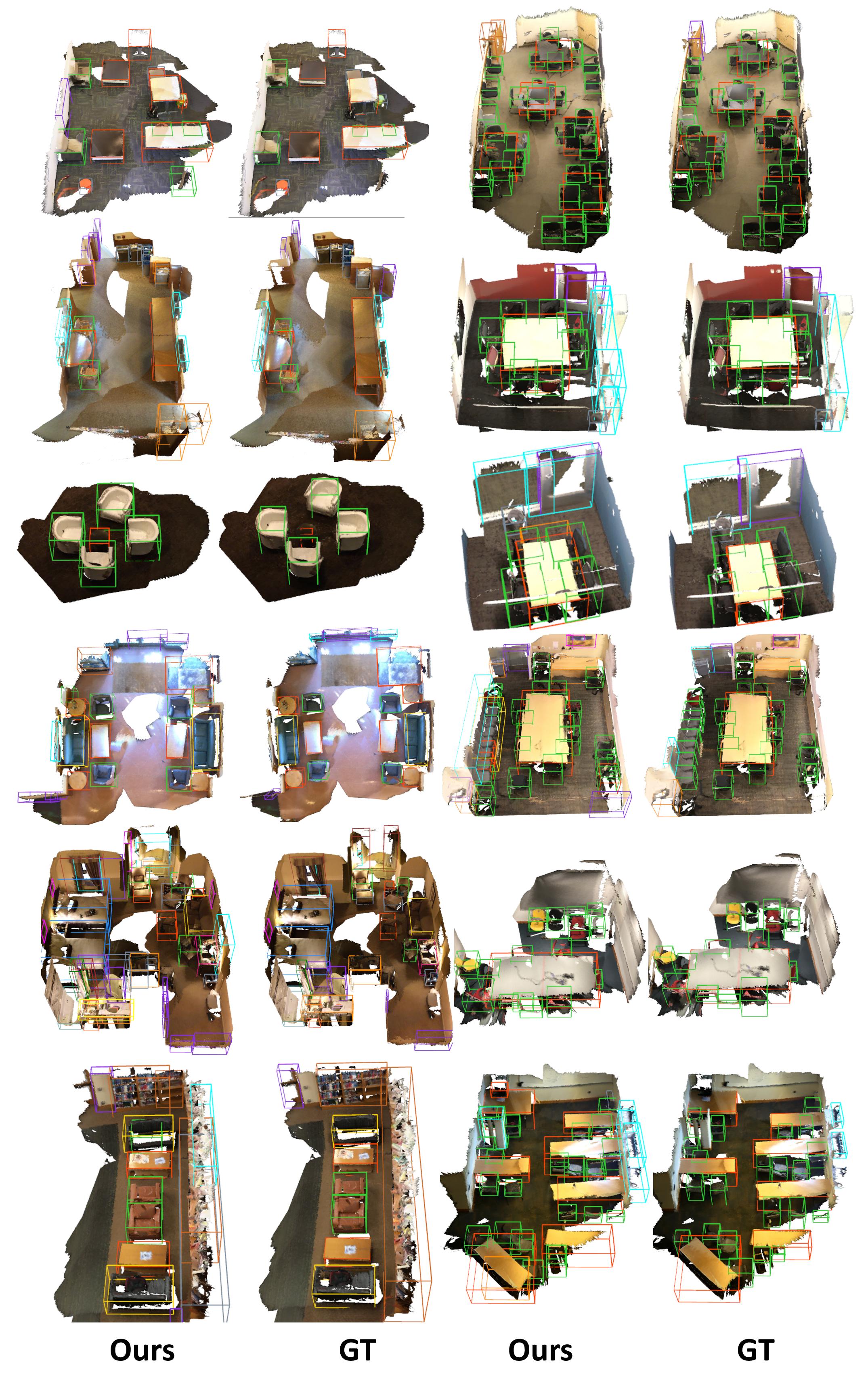}
\caption{\small{More qualitative results on ScanNet V2.}}
\label{Figure:Qual:Scan}
\end{figure}

\begin{figure}
\centering
\includegraphics[width=0.9\textwidth]{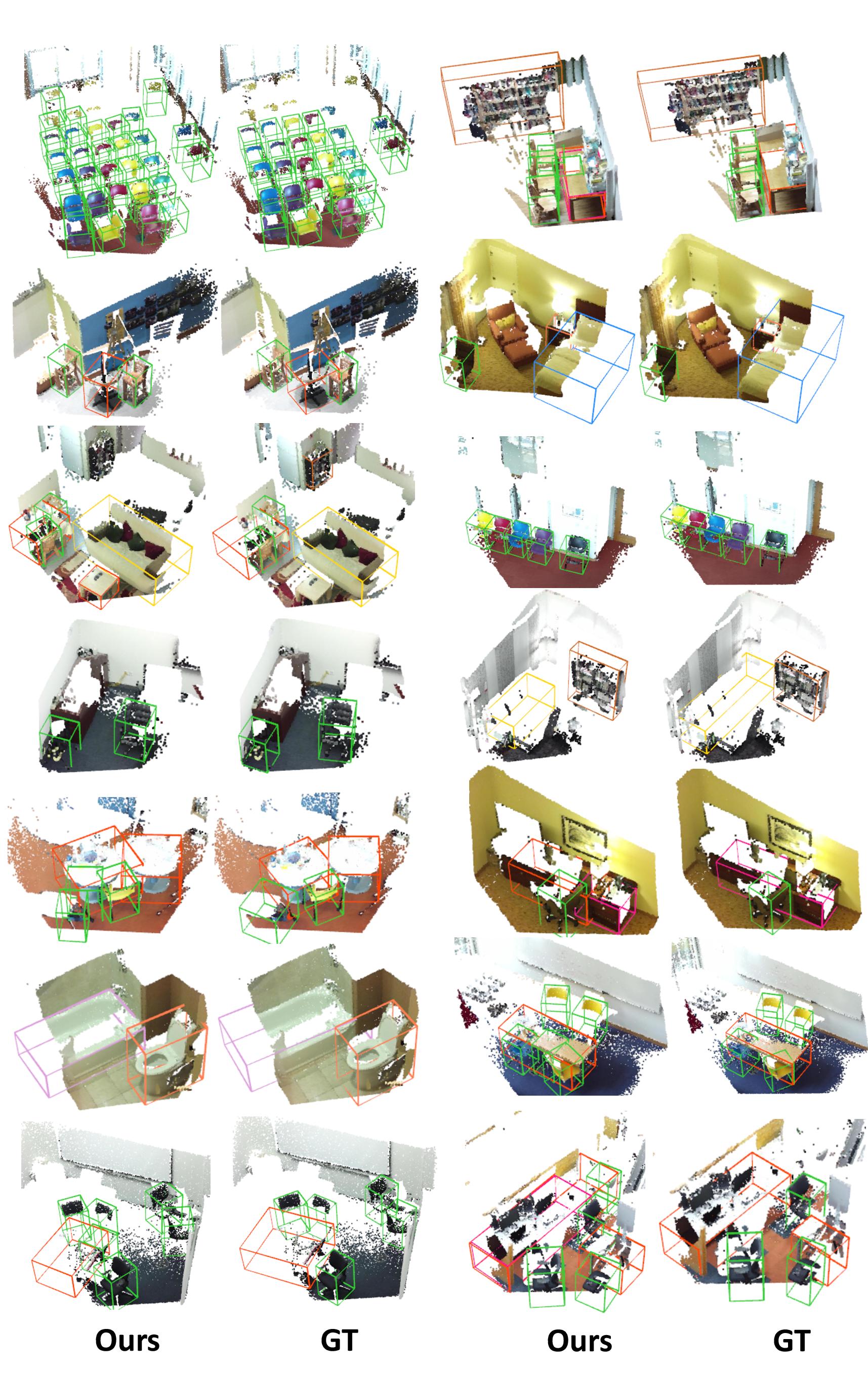}
\caption{\small{More qualitative results on SUN RGB-D V1.}}
\label{Figure:Qual:SUN}
\end{figure}

We show more qualitative examples of 3D object detection for both datasets in Figure \ref{Figure:Qual:Scan} and \ref{Figure:Qual:SUN}. In Figure \ref{Figure:Qual:thin}, we show qualitative comparisons between our approach and the top-performing baseline approach on thin objects. Our method is more accurate and detects more positive thin objects than baseline approaches. We also show some failure cases with our approach in Figure \ref{Figure:Qual:failures}, and we summarize the failure patterns in the caption.

